\documentclass{article}
\usepackage[utf8]{inputenc}
\usepackage[dvipsnames,rgb,dvips]{xcolor}
\usepackage{graphicx}
\usepackage{float}
\usepackage{epsfig}
\usepackage{graphicx}
\usepackage{psfrag}
\usepackage{dcolumn}
\usepackage{bm}
\usepackage{amsmath}
\usepackage{amssymb}
\usepackage{acronym}
\usepackage[rflt]{floatflt}
\usepackage{latexsym}
\usepackage{listings}
\usepackage{verbatim}
\usepackage{subcaption}
\usepackage{hyperref}
\usepackage{xcolor}
\usepackage{comment}
\usepackage{arydshln}

\newcommand{\Cov}[2]{\text{Cov}_{#1}\left[#2\right]}

\newcommand{\bphi}{\mathbf{\phi}}

\newcommand{\beps}{\mathbf{\epsilon}}

\newcommand{\D}{\mathcal{D}}

\title{A Comprehensive guide to Bayesian Convolutional Neural Network with Variational Inference}
 
\author{
    Kumar Shridhar\textsuperscript{1,2,3},
    Felix Laumann\textsuperscript{3,4},
    Marcus Liwicki\textsuperscript{1,5}\\
    \\
    \textsuperscript{1} MindGarage, 
    \textsuperscript{2} Technical University Kaiserslautern\\
    \textsuperscript{3} NeuralSpace,
    \textsuperscript{4} Imperial College London\\
    \textsuperscript{5} Lule\aa \ University of Technology\\}

\date{December 2018}

\begin{document}

\maketitle

\begin{abstract}
Artificial Neural Networks are connectionist systems that perform a given task by learning on examples without having prior knowledge about the task. This is done by finding an optimal point estimate for the weights in every node.
Generally, the network using point estimates as weights perform well with large datasets, but they fail to express uncertainty in regions with little or no data, leading to overconfident decisions.
\newline
In this paper, Bayesian Convolutional Neural Network (BayesCNN) using Variational Inference is proposed, that introduces probability distribution over the weights. Furthermore, the proposed BayesCNN architecture is applied to tasks like Image Classification, Image Super-Resolution and Generative Adversarial Networks. The results are compared to point-estimates based architectures on MNIST, CIFAR-10 and CIFAR-100 datasets for Image CLassification task, on BSD300 dataset for Image Super Resolution task and on CIFAR10 dataset again for Generative Adversarial Network task. 

BayesCNN is based on Bayes by Backprop which derives a variational approximation to the true posterior.  We, therefore, introduce the idea of applying two convolutional operations, one for the mean and one for the variance.
Our proposed method not only achieves performances equivalent to frequentist inference in identical architectures but also incorporate a measurement for uncertainties and regularisation. It further eliminates the use of dropout in the model. Moreover, we predict how certain the model prediction is based on the epistemic and aleatoric uncertainties and empirically show how the uncertainty can decrease, allowing the decisions made by the network to become more deterministic as the training accuracy increases.
Finally, we propose ways to prune the Bayesian architecture and to make it more computational and time effective. 
\newline

\end{abstract}

\section{Introduction}
\newacro{cnn}[\textsc{Cnn}]{Convolutional neural network}

Deep Neural Networks (DNNs), are connectionist systems that learn to perform tasks by learning on examples without having prior knowledge about the tasks. 
They easily scale to millions of data points and yet remain tractable to optimize with stochastic gradient descent.

\acp{cnn}, a variant of DNNs, have already surpassed human accuracy in the realm of image classification (e.g. \cite{he2016deep}, \cite{simonyan2014very}, \cite{krizhevsky2012imagenet}). Due to the capacity of \acp{cnn} to fit on a wide diversity of non-linear data points, they require a large amount of training data. This often makes \acp{cnn} and Neural Networks in general, prone to overfitting on small datasets. The model tends to fit well to the training data, but are not predictive for new data. This often makes the Neural Networks incapable of correctly assessing the uncertainty in the training data and hence leads to overly confident decisions about the correct class, prediction or action.

Various regularization techniques for controlling over-fitting are used in practice namely early stopping, weight decay, L1, L2 regularizations and currently the most popular and empirically effective technique being \emph{dropout}~\cite{hinton2012improving}.

\subsection{Problem Statement}

Despite Neural Networks architectures achieving state-of-the-art results in almost all classification tasks, Neural Networks still make over-confident decisions. A measure of uncertainty in the prediction is missing from the current Neural Networks architectures. Very careful training, weight control measures like regularization of weights and similar techniques are needed to make the models susceptible to over-fitting issues. 

We address both of these concerns by introducing Bayesian learning to Convolutional Neural Networks that adds a measure of uncertainty and regularization in their predictions. 

\subsection{Current Situation}

Deep Neural Networks have been successfully applied to many domains, including very sensitive domains like health-care, security, fraudulent transactions and many more. However, from a probability theory perspective, it is unjustifiable to use single point-estimates as weights to base any classification on.
On the other hand, Bayesian neural networks are more robust to over-fitting, and can easily learn from small datasets. The Bayesian approach further offers uncertainty estimates via its parameters in form of probability distributions (see Figure 1.1). At the same time, by using a prior probability distribution to integrate out the parameters, the average is computed across many models during training, which gives a regularization effect to the network, thus preventing overfitting.

Bayesian posterior inference over the neural network parameters is a theoretically attractive method for controlling overfitting; however, modelling a distribution over the kernels (also known as filters) of a \acp{cnn} has never been attempted successfully before, perhaps because of the vast number of parameters and extremely large models commonly used in practical applications.

Even with a small number of parameters, inferring model posterior in a Bayesian NN is a difficult task. Approximations to the model posterior are often used instead, with the variational inference being a popular approach. In this approach one would model the posterior using a simple \textit{variational} distribution such as a Gaussian, and try to fit the distribution's parameters to be as close as possible to the true posterior. This is done by minimising the \textit{Kullback-Leibler divergence} from the true posterior. Many have followed this approach in the past for standard NN models \cite{hinton1993keeping}, \cite{barber1998ensemble}, \cite{graves2011practical}, \cite{blundell2015weight}.
But the variational approach used to approximate the posterior in Bayesian NNs can be fairly computationally expensive -- the use of Gaussian approximating distributions increases the number of model parameters considerably, without increasing model capacity by much. \cite{blundell2015weight} for example used Gaussian distributions for Bayesian NN posterior approximation and have doubled the number of model parameters, yet report the same predictive performance as traditional approaches using dropout. This makes the approach unsuitable in practice to use with \acp{cnn} as the increase in the number of parameters is too costly.

\subsection{Our Hypothesis}
We build our Bayesian \ac{cnn} upon \textit{Bayes by Backprop} \cite{graves2011practical}, \cite{blundell2015weight}. The exact Bayesian inference on the weights of a neural network is intractable as the number of parameters is very large and the functional form of a neural network does not lend itself to exact integration. So, we approximate the intractable true posterior probability distributions $p(w|\mathcal{D})$ with variational probability distributions $q_{\theta}(w|\mathcal{D})$, which comprise the properties of Gaussian distributions $\mu \in \mathbb{R}^d$ and $\sigma \in \mathbb{R}^d$, denoted by $\mathcal{N}(\theta|\mu, \sigma^2)$, where $d$ is the total number of parameters defining a probability distribution. The shape of these Gaussian variational posterior probability distributions, determined by their variance $\sigma^2$, expresses an uncertainty estimation of every model parameter. \\ \\

\begin{figure}[H]
\begin{center}
\includegraphics[height=.28\textheight]{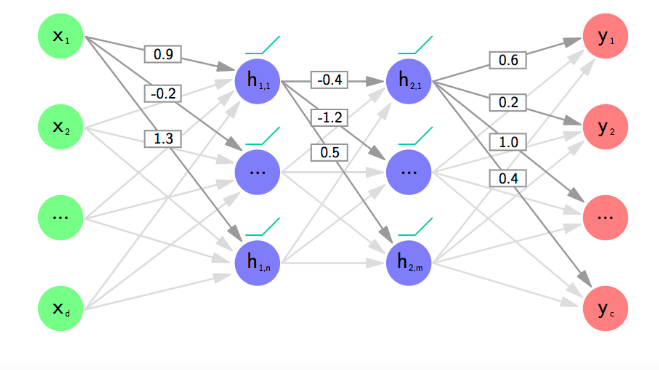}
\includegraphics[height=.28\textheight]{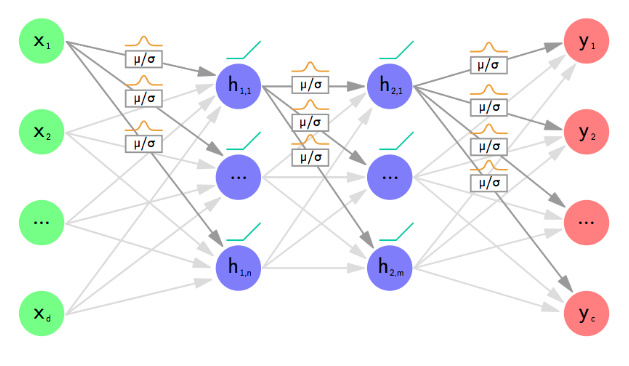}
\label{fig:Scalar_Bayesian_Distribution_Gluon}
\caption{Top: Each filter weight has a fixed value, as in the case of frequentist Convolutional Networks. Bottom: Each filter weight has a distribution, as in the case of Bayesian Convolutional Networks. \cite{Gluon}}
\end{center}
\end{figure}

\subsection{Our Contribution}
The main contributions of our work are as follows: 
\begin{enumerate}
    \item We present how \textit{Bayes by Backprop} can be efficiently applied to \acp{cnn}. We, therefore, introduce the idea of applying two convolutional operations, one for the mean and one for the variance.
    \item We show how the model learns richer representations and predictions from cheap model averaging.
    \item We empirically show that our proposed generic and reliable variational inference method for Bayesian \acp{cnn} can be applied to various \ac{cnn} architectures without any limitations on their performances. 
    \item We examine how to estimate the aleatoric and epistemic uncertainties and empirically show how the uncertainty can decrease, allowing the decisions made by the network to become more deterministic as the training accuracy increases. 
    \item We also empirically show how our method typically only doubles the number of parameters yet trains an infinite ensemble using unbiased Monte Carlo estimates of the gradients. 
    \item We also apply L1 norm to the trained model parameters and prune the number of non zero values and further, fine-tune the model to reduce the number of model parameters without a reduction in the model prediction accuracy. 
    \item Finally, we will apply the concept of Bayesian CNN to tasks like Image Super-Resolution and Generative Adversarial Networks and we will compare the results with other prominent architectures in the respective domain.
\end{enumerate} 
This work builds on the foundations laid out by Blundell et al. \cite{blundell2015weight}, who introduced \textit{Bayes by Backprop} for feedforward neural networks. Together with the extension to recurrent neural networks, introduced by Fortunato et al. \cite{fortunato2017bayesian}, \textit{Bayes by Backprop} is now applicable on the three most frequently used types of neural networks, i.e., feedforward, recurrent, and convolutional neural networks.

\section{Background}

\subsection{Neural Networks}
\subsubsection{Brain Analogies}

A perceptron is conceived as a mathematical model of how the neurons function in our brain by a famous psychologist Rosenblatt. According to Rosenblatt, a neuron  takes a set of binary inputs (nearby neurons), multiplies each input by a continuous-valued weight (the synapse strength to each nearby neuron), and thresholds the sum of these weighted inputs to output a 1 if the sum is big enough and otherwise a 0 (the same way neurons either fire or do not fire).

\begin{figure}[H]
\begin{center}
\includegraphics[height=.28\textheight]{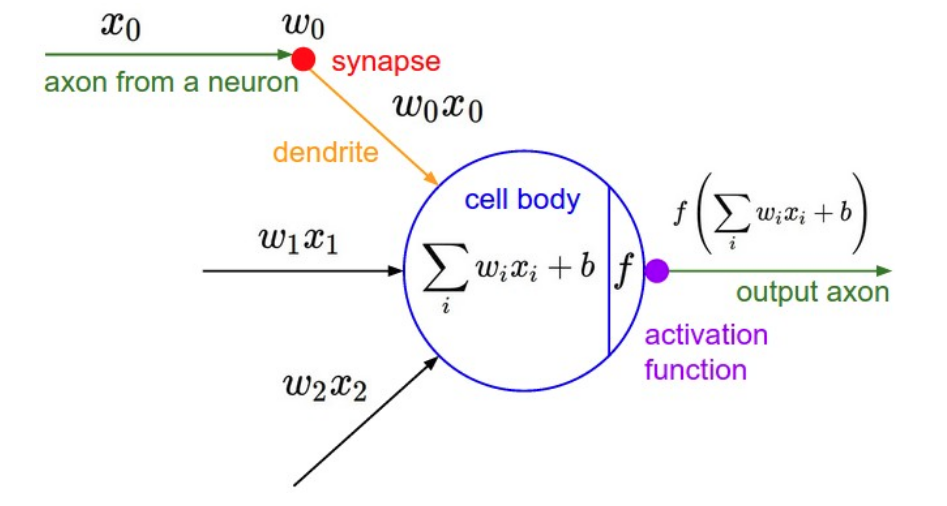}
\label{fig:Neural_Network}
\caption{Biologically inspired Neural Network \cite{karparthy}}
\end{center}
\end{figure}

\subsubsection{Neural Network}

Inspired by the biological nervous system, the structure of an Artificial Neural Network (ANN) was developed to process information similar to how brain process information. A large number of highly interconnected processing elements (neurons) working together makes a Neural Network solve complex problems. Just like humans learn by example, so does a Neural Network. Learning in biological systems involves adjustments to the synaptic connections which is similar to weight updates in a Neural Network. 

A Neural Network consists of three layers: input layer to feed the data to the model to learn representation, hidden layer that learns the representation and the output layer that outputs the results or predictions. Neural Networks can be thought of an end to end system that finds patterns in data which are too complex to be recognized by a human to teach to a machine.

\begin{figure}[H]
\begin{center}
\includegraphics[height=.38\textheight]{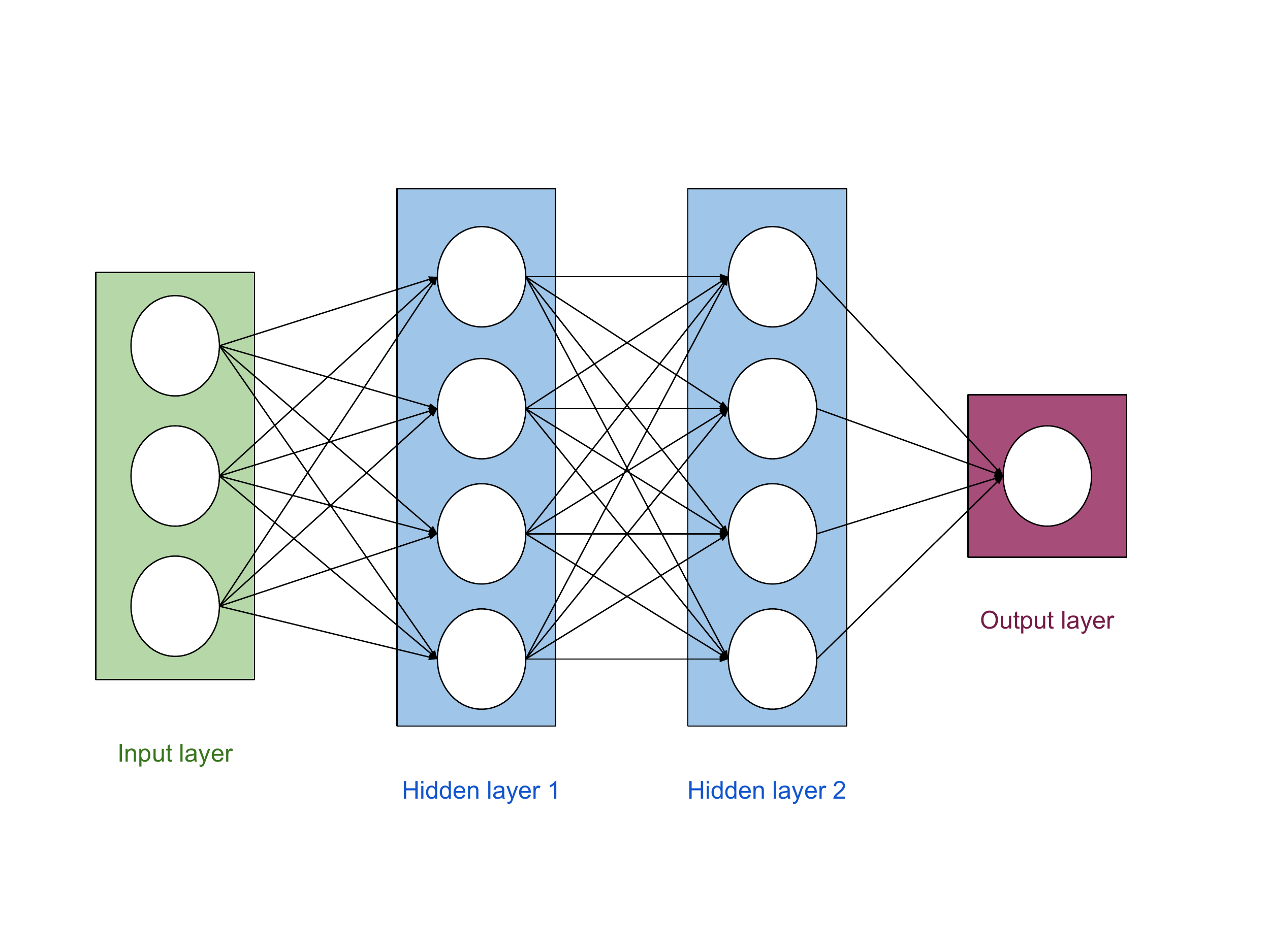}
\label{fig:Two Layered Neural_Network}
\caption{Neural Network with two hidden layers}
\end{center}
\end{figure}

\subsubsection{Convolutional Neural Network}

Hubel and Wiesel in their hierarchy model mentioned a neural network to have a hierarchy structure in the visual cortex. LGB (lateral geniculate body) forms the simple cells that form the complex cells which form the lower order hypercomplex cells that finally form the higher order hypercomplex cells.
Also, the network between the lower order hypercomplex cells and the higher order hypercomplex cells are structurally similar to the network between simple cells and the complex cells. In this hierarchy, a cell in a higher stage generally has a tendency to respond selectively to a more complicated feature of the stimulus pattern, and the cell at the lower stage responds to simpler features. Also, higher stage cells possess a larger receptive field and are more insensitive to the shift in the position of the stimulus pattern.

Similar to a hierarchy model, a neural network starting layers learns simpler features like edges and corners and subsequent layers learn complex features like colours, textures and so on. Also, higher neural units possess a larger receptive field which builds over the initial layers. However, unlike in multilayer perceptron where all neurons from one layer are connected with all the neurons in the next layer, weight sharing is the main idea behind a convolutional neural network. 
Example: instead of each neuron having a different weight for each pixel of the input image (28*28 weights), the neurons only have a small set of weights (5*5) that is applied to a whole bunch of small subsets of the image of the same size. Layers past the first layer work in a similar way by taking in the ‘local’ features found in the previously hidden layer rather than pixel images, and successively see larger portions of the image since they are combining information about increasingly larger subsets of the image. Finally, the final layer makes the correct prediction for the output class.

The reason for why this is helpful is intuitive if not mathematically clear: without such constraints, the network would have to learn the same simple things (such as detecting edges, corners, etc) a whole bunch of times for each portion of the image. But with the constraint there, only one neuron would need to learn each simple feature - and with far fewer weights overall, it could do so much faster! Moreover, since the pixel-exact locations of such features do not matter the neuron could basically skip neighbouring subsets of the image - subsampling, now known as a type of pooling - when applying the weights, further reducing the training time. The addition of these two types of layers - convolutional and pooling layers - are the primary distinctions of Convolutional Neural Nets (CNNs/ConvNets) from plain old neural nets.

\subsection{Probabilistic Machine Learning}
\subsubsection{Variational Inference}

We define a function $y = f(\mathbf{x})$ that estimates the given inputs $\{ x_1, \hdots, x_N \}$ and their corresponding outputs $\{y_1, \hdots, y_N\}$ and produces an predictive output. Using Bayesian inference, a prior distribution is used over the space of functions $p(f)$. This distribution represents our prior belief as to which functions are likely to have generated our data. 

A likelihood is defined as $p(Y | f, X)$ to capture the process in which given a function observation is generated. We use the Bayes rule to find the posterior distribution given our dataset: $p(f | X, Y)$.

The new output can be predicted for a new input point $x^*$ by integrating over all possible functions $f$,
\begin{align} \label{eq:post}
p(y^* | x^*, X, Y) = \int p(y^* | f^*) p(f^* | x^*, X, Y) df^*
\end{align}

The equation \eqref{eq:post} is intractable due to the integration sign. We can approximate it by taking a finite set of random variables $w$ and conditioning the model on it. However, it is based on a modelling assumption that the model depends on these variables alone, and making them into sufficient statistics in our approximate model.

The predictive distribution for a new input point $x^*$ is then given by 
$$
p(y^* | x^*, X, Y) = \int p(y^* | f^*) p(f^* | x^*, w) p(w | X, Y) df^* d w.
$$

However, the distribution $p(w | X, Y)$ still remains intractable and we need to approximate it with a variational distribution $q(w)$, that can be computed. The approximate distribution needs to be as close as possible to the posterior distribution obtained from the original model. We thus minimise the Kullback--Leibler (KL) divergence, intuitively a measure of similarity between two distributions: $KL(q(w) ~||~ p(w | X, Y))$,
resulting in the approximate predictive distribution 
\begin{align} \label{eq:predictive}
q(y^* | x^*) = \int p(y^* | f^*) p(f^* | x^*, w) q(w)  df^* dw.
\end{align}

Minimising the Kullback--Leibler divergence is equivalent to maximising the \textit{log evidence lower bound},

\begin{align} \label{eq:L:VI}
KL_{\text{VI}} := \int q(w) p(F | X, w) \log p(Y | F) dF dw - KL(q(w) || p(w)) 
\end{align}

with respect to the variational parameters defining $q(w)$. This is known as \textit{variational inference}, a standard technique in Bayesian modelling.

Maximizing the KL divergence between the posterior and the prior over $w$ will result in a variational distribution that learns a good representation from the data (obtained from log likelihood) and is closer to the prior distribution. In other words, it can prevent overfitting.

\subsubsection{Local  Reparametrisation  Trick}
The ability to rewrite statistical problems in an equivalent but different form, to reparameterise them, is one of the most general-purpose tools used in mathematical statistics. The type of \textit{reparameterization} when the global uncertainty in the weights is translated into a form of local uncertainty which is independent across examples is known as the \emph{local reparameterization trick}. 
An alternative estimator is deployed for which $\Cov{}{L_{i},L_{j}} = 0$, so that the variance of the stochastic gradients scales as $1/M$.
The new estimator is made computationally efficient by sampling the intermediate variables and not sampling $\beps$ directly, but only  $f(\beps)$ through which $\beps$ influences $L_\D^\text{SGVB}(\bphi)$. Hence, the source of global noise can be translated to local noise ($\beps \rightarrow f(\beps)$), a local reparameterization can be applied so as to obtain a statistically efficient gradient estimator.

The technique can be explained through a simple example: We consider an input($X$) of random uniform function ranging from -1 to +1 and an output($Y$) as a random normal distribution around mean $X$ and standard deviation $\delta$. The Mean Squared Loss would be defined as 
$(Y-X)^2$. The problem here is during the backpropagation from the random normal distribution function. As we are trying to propagate through a stochastic node we reparameterize by adding $X$ to the random normal function output and multiplying by $\delta$.
The movement of parameters out of the normal distribution does not change the behaviour of the model.

\subsection{Uncertainties in Bayesian Learning}

Uncertainties in a network is a measure of how certain the model is with its prediction. In Bayesian modelling, there are two main types of uncertainty one can model \cite{Kiureghian}: \textit{Aleatoric} uncertainty and \textit{Epistemic} uncertainty. 

\textit{Aleatoric} uncertainty measures the noise inherent in the observations. This type of uncertainty is present in the data collection method like the sensor noise or motion noise which is uniform along the dataset. This cannot be reduced if more data is collected. \textit{Epistemic} uncertainty, on the other hand, represents the uncertainty caused by the model. This uncertainty can be reduced given more data and is often referred to as \textit{model uncertainty}.  Aleatoric uncertainty can further be categorized into \textit{homoscedastic} uncertainty, uncertainty which stays constant for different inputs, and \textit{heteroscedastic} uncertainty. Heteroscedastic uncertainty depends on the inputs to the model, with some inputs potentially having more noisy outputs than others. Heteroscedastic uncertainty is in particular important so that model prevents from outputting very confident decisions.

Current work measures uncertainties by placing a probability distributions over either the model parameters or model outputs. Epistemic uncertainty is modelled by placing a prior distribution over a model's weights and then trying to capture how much these weights vary given some data. Aleatoric uncertainty, on the other hand, is modelled by placing a distribution over the output of the model.

\subsubsection{Sources of Uncertainities}

The following can be the source of uncertainty as mentioned by Kiureghian \cite{Kiureghian}:

\begin{enumerate}
    \item Uncertainty inherent in the basic random variables $X$, such as the uncertainty inherent in material property constants and load values, which can be directly measured.
    \item Uncertain model error resulting from the selection of the form of the probabilistic sub-model $f_{X}(x,H_{f})$ used to describe the distribution of basic variables.
    \item  Uncertain modeling errors resulting from selection of the physical sub-models $g_{i}(x,H_{g}), i = 1,2,...,m,$ used to describe the derived variables.
    \item  Statistical uncertainty in the estimation of the parameters $H_f$ of the probabilistic sub-model.
    \item  Statistical uncertainty in the estimation of the parameters $H_g$ of the physical sub-models.
    \item  Uncertain errors involved in measuring of observations, based on which the parameters $H_f$ and $H_g$  are estimated. These include errors involved in indirect measurement, e.g., the measurement of a quantity through a proxy, as in non-destructive testing of material strength.
    \item Uncertainty modelled by the random variables $Y$ corresponding to the derived variables $y$, which may include, in addition to all the above uncertainties, uncertain errors resulting from computational errors, numerical approximations or truncations. For example, the computation of load effects in a nonlinear structure by a finite element procedure employs iterative calculations, which invariably involve convergence tolerances and truncation errors.
\end{enumerate} 

\subsection{Backpropagation}

Backpropagation in a Neural Networks was proposed by Rumelhart \cite{Rumelhart} in 1986 and it is the most commonly used method for training neural networks. Backpropagation is a technique to compute the gradient of the loss in terms of the network weights. It operates in two phases: firstly, the input features through the network propagates in the forward direction to compute the function output and thereby the loss associated with the parameters. Secondly, the derivatives of the training loss with respect to the weights
are propagated back from the output layer towards the input layers.
These computed derivatives are further used to update the weights of the network. This is a continuous process and updating of the weight occurs continuously over every iteration. 

Despite the popularity of backpropagation, there are many hyperparameters in backpropagation based stochastic optimization that requires specific tuning, e.g., learning rate, momentum, weight decay, etc. The time required for finding the optimal values is proportional to the data size. For a network trained with backpropagation, only point estimates of the weights are achieved in the network. As a result, these networks make overconfident predictions and do not account for uncertainty in the parameters. Lack of uncertainty measure makes the network prone to overfitting and a need for regularization.

A Bayesian approach to Neural Networks provides the shortcomings with the backpropagation approach \cite{mackay1996hyperparameters} as Bayesian methods naturally account for uncertainty in parameter estimates and can propagate this uncertainty into predictions.
Also, averaging over parameter values instead of just choosing single point estimates makes the model robust to overfitting. 

Sevreal approaches has been proposed in the past for learning in Bayesian Networks: Laplace approximation \cite{Mackay1991APB}, MC Dropout \cite{gal2015bayesian},and Variational Inference \cite{hinton1993keeping} \cite{graves2011practical} \cite{blundell2015weight}. We used Bayes by Backprop \cite{blundell2015weight} for our work and is explained next.

\subsubsection{Bayes by Backprop}
\textit{Bayes by Backprop} \cite{graves2011practical, blundell2015weight} is a variational inference method to learn the posterior distribution on the weights $w \sim q_{\theta}(w|\mathcal{D})$ of a neural network from which weights $w$ can be sampled in backpropagation. 
It regularises the weights by minimising a compression cost, known as the variational free energy or the expected lower bound on the marginal likelihood.

Since the true posterior is typically intractable, an approximate distribution $q_{\theta}(w|\mathcal{D})$ is defined that is aimed to be as similar as possible to the true posterior $p(w|\mathcal{D})$, measured by the Kullback-Leibler (KL) divergence \cite{kullback1951information}. Hence, we define the optimal parameters $\theta^{opt}$ as
\begin{equation}
    \begin{aligned} \label{KL}
        \theta^{opt}&=\underset{\theta}{\text{arg min}}\ \text{KL} \ [q_{\theta}(w|\mathcal{D})\|p(w|\mathcal{D})] \\
        &=\underset{\theta}{\text{arg min}}\ \text{KL} \ [q_{\theta}(w|\mathcal{D})\|p(w)] \\ & -\mathbb{E}_{q(w|\theta)}[\log p(\mathcal{D}|w)]+\log p(\mathcal{D})
    \end{aligned}
\end{equation}

where
\begin{equation}
    \text{KL} \ [q_{\theta}(w|\mathcal{D})\|p(w)]= \int q_{\theta}(w|\mathcal{D})\log\frac{q_{\theta}(w|\mathcal{D})}{p(w)}dw .
\end{equation}
This derivation forms an optimisation problem with a resulting cost function widely known as \textit{variational free energy} \cite{neal1998view,yedidia2005constructing,friston2007variational} which is built upon two terms: the former, $\text{KL} \ [q_{\theta}(w|\mathcal{D})\|p(w)]$, is dependent on the definition of the prior $p(w)$, thus called complexity cost, whereas the latter, $\mathbb{E}_{q(w|\theta)}[\log p(\mathcal{D}|w)]$, is dependent on the data $p(\mathcal{D}|w)$, thus called likelihood cost. 
The term $\log p(\mathcal{D})$ can be omitted in the optimisation because it is constant.
\newline Since the KL-divergence is also intractable to compute exactly, we follow a stochastic variational method \cite{graves2011practical,blundell2015weight}.
We sample the weights $w$ from the variational distribution $q_{\theta}(w|\mathcal{D})$ since it is much more probable to draw samples which are appropriate for numerical methods from the variational posterior $q_{\theta}(w|\mathcal{D})$ than from the true posterior $p(w|\mathcal{D})$. Consequently, we arrive at the tractable cost function \eqref{cost} which is aimed to be optimized, i.e. minimised w.r.t. $\theta$, during training:
\begin{equation} \label{cost}
    \mathcal{F}(\mathcal{D}, \theta)\approx \sum_{i=1}^n \log q_{\theta}(w^{(i)}|\mathcal{D})-\log p(w^{(i)})-\log p(\mathcal{D}|w^{(i)})
\end{equation}
where $n$ is the number of draws.
\newline We sample $w^{(i)}$ from $q_{\theta}(w|\mathcal{D})$. The uncertainty afforded by \textit{Bayes by Backprop} trained neural networks has been used successfully for training feedforward neural networks in both supervised and reinforcement learning environments \cite{blundell2015weight,lipton2016efficient,houthooft2016curiosity}, for training recurrent neural networks \cite{fortunato2017bayesian}, but has not been applied to convolutional neural networks to-date.

\subsection{Model Weights Pruning}

Model pruning reduces the sparsity in a deep neural network's
various connection matrices, thereby reducing the number of valued parameters in the model. The whole idea of model pruning is to reduce the number of parameters without much loss in the accuracy of the model. This reduces the use of a large parameterized model with regularization and promotes the use of dense connected smaller models. Some recent work suggests \cite{DBLP:journals/corr/HanMD15} \cite{DBLP:journals/corr/NarangDSE17} that the network can achieve a sizable reduction in model size, yet achieving comparable accuracy. 
Model pruning possesses several advantages in terms of reduction in computational cost, inference time and in energy-efficiency. The resulting pruned model typically has sparse connection matrices, so efficient inference using these sparse models requires purpose-built hardware capable of loading sparse matrices and/or performing sparse matrix-vector operations. Thus the overall memory usage is reduced with the new pruned model.

There are several ways of achieving the pruned model, the most popular one is to map the low contributing weights to zero and reducing the number of overall non-zero valued weights. This can be achieved by training a large sparse model and pruning it further which makes it comparable to training a small dense model. 

Assigning weights zero to most features and non-zero weights to only important features can be formalized by applying the $L_0$ norm, where $L_0 = ||\theta||_0 = \sum{_j} \delta (\theta_j \neq 0)$, and it applies a constant penalty to all non-zero weights. 
$L_0$ norm can be thought of a feature selector norm that only assigns non-zero values to feature that are important. However, the $L_0$ norm is non-convex and hence, non-differentiable that makes it a NP-hard problem and can be only efficiently solved when $P = NP$.
The alternative that we use in our work is the $L_1$ norm, which is equal to the sum of the absolute weight values, $||\theta||_1 = \sum_j |\theta_j|$. $L_1$ norm is convex and hence differentiable and can be used as an approximation to $L_0$ norm  \cite{tibshirani1996regression}. $L_1$ norm works as a sparsity inducing regularizer by making large number of coefficients equal to zero, working as a great feature selector in our case. Only thing to keep in mind is that the $L_1$ norm do not have a gradient at $\theta_j = 0$ and we need to keep that in mind. 

Pruning away the less salient features to zero has been used in this work and is explained in details in Our Contribution section. 

\section{Related Work}

\subsection{Bayesian Training}
Applying Bayesian methods to neural networks has been studied in the past with various approximation methods for the intractable true posterior probability distribution $p(w|\mathcal{D})$. Buntine and Weigend \cite{buntine1991bayesian} started to propose various \textit{maximum-a-posteriori} (MAP) schemes for neural networks. They were also the first who suggested second order derivatives in the prior probability distribution $p(w)$ to encourage smoothness of the resulting approximate posterior probability distribution.
In subsequent work by Hinton and Van Camp \cite{hinton1993keeping}, the first variational methods were proposed which naturally served as a regularizer in neural networks. He also mentioned that the amount of information in weight can be controlled by adding Gaussian noise. When optimizing the trade-off between the expected error and the information in the weights, the noise level can be adapted during learning.

Hochreiter and Schmidhuber \cite{hochreiter1995simplifying} suggested taking an information theory perspective into account and utilising a minimum description length (MDL) loss. This penalises non-robust weights by means of an approximate penalty based upon perturbations of the weights on the outputs.
Denker and LeCun \cite{denker1991transforming}, and MacKay \cite{mackay1995probable} investigated the posterior probability distributions of neural networks and treated the search in the model space (the space of architectures, weight decay, regularizers, etc..) as an inference problem and tried to solve it using Laplace approximations.
As a response to the limitations of Laplace approximations, Neal \cite{neal2012bayesian} investigated the use of hybrid Monte Carlo for training neural networks, although it has so far been difficult to apply these to the large sizes of neural networks built in modern applications. 
Also, these approaches lacked scalability in terms of both the network architecture and the data size.

More recently, Graves \cite{graves2011practical} derived a variational inference scheme for neural networks and Blundell et al. \cite{blundell2015weight} extended this with an update for the variance that is unbiased and simpler to compute. Graves \cite{graves2016stochastic} derives a similar algorithm in the case of a mixture posterior probability distribution. 
A more scalable solution based on expectation propagation was proposed by Soudry \cite{Soudry:NIPS2014_5269} in 2014. While this method works for networks with binary weights, its extension to continuous weights is unsatisfying as it does not produce estimates of posterior variance.

Several authors have claimed that Dropout \cite{srivastava2014dropout} and Gaussian Dropout \cite{wang2013fast} can be viewed as approximate variational inference schemes \cite{gal2015bayesian, kingma2015variational}. We compare our results to Gal's \& Ghahramani's \cite{gal2015bayesian} and discuss the methodological differences in detail.

\subsection{Uncertainty Estimation}

Neural Networks can predict uncertainty when Bayesian methods are introduced in it. An attempt to model uncertainty has been studied from 1990s \cite{neal2012bayesian} but has not been applied successfully until 2015. Gal and Ghahramani \cite{Gal2015Dropout} in 2015 provided a theoretical framework for modelling Bayesian uncertainty. Gal and Ghahramani \cite{gal2015bayesian} obtained the uncertainty estimates by casting dropout training in conventional deep networks as a Bayesian approximation of a Gaussian Process. They showed that any network trained with dropout is an approximate Bayesian model, and uncertainty estimates can be obtained by computing the variance on multiple predictions with different dropout masks.

\subsection{Model Pruning}

Some early work in the model pruning domain used a second-order Taylor approximation of the increase in the loss function of the network when weight is set to zero \cite{lecun1990optimal}. A diagonal Hessian approximation was used to calculate the saliency for each parameter \cite{lecun1990optimal} and the low-saliency parameters were pruned from the network and the network was retrained.

Narang \cite{DBLP:journals/corr/NarangDSE17} showed that a pruned RNN and GRU model performed better for the task of speech recognition compared to a dense network of the original size. This result is very similar to the results obtained in our case where a pruned model achieved better results than a normal network. However, no comparisons can be drawn as the model architecture (CNN vs RNN) used and the tasks (Computer Vision vs Speech Recognition) are completely different.  Narang \cite{DBLP:journals/corr/NarangDSE17} in his work introduced a gradual pruning scheme based on pruning all the weights in a layer
less than some threshold (manually chosen) which is linear with some
slope in phase 1 and linear with some slope in phase 2 followed by
normal training. However, we reduced the number of filters to half for one case and in the other case, we induced a sparsity-based on L1 regularization to remove the less contributing weights and reduced the parameters.

Other similar work \cite{DBLP:journals/corr/AnwarHS15, DBLP:journals/corr/LebedevGROL14, DBLP:journals/corr/ChangpinyoSZ17} to our work that reduces or removed the redundant connections or induces sparsity are motivated by the desire to speed up computation.
The techniques used are highly convolutional layer dependent and is not applicable to other architectures like Recurrent Neural Networks. 
One another interesting method of pruning is to represent each parameter with a smaller floating point number like 16-bits instead of 64 bits. This way there is a speed up in the training and inference time and the model is less computationally expensive. 

Another viewpoint for model compression was presented by Gong \cite{DBLP:journals/corr/GongLYB14}. He proposed vector quantization to achieve different compression ratios and different accuracies and depending on the use case, the compression and accuracies can be chosen. However, it requires a different hardware architecture to support the inference at runtime. Besides quantization, other potentially complementary approaches to reducing model size include low-rank matrix factorization \cite{denton2014exploiting, DBLP:journals/corr/LebedevGROL14} and group sparsity regularization to arrive at an optimal layer size \cite{DBLP:conf/nips/AlvarezS16}.

\section{Our Concept}

\subsection{Bayesian Convolutional Neural Networks with Variational Inference}
In this section, we explain our algorithm of building a \ac{cnn} with probability distributions over its weights in each filter, as seen in Figure \ref{fig:filter_scalar}, and apply variational inference, i.e. \textit{Bayes by Backprop}, to compute the intractable true posterior probability distribution, as described in the last Chapter. Notably, a fully Bayesian perspective on a \ac{cnn} is for most \ac{cnn} architectures not accomplished by merely placing probability distributions over weights in convolutional layers; it also requires probability distributions over weights in fully-connected layers (see Figure \ref{fig:CNNwithdist_grey}). 
\begin{figure}[H] 
\centering
\begin{minipage}{.4\textwidth}
\centering
\includegraphics[width=\linewidth]{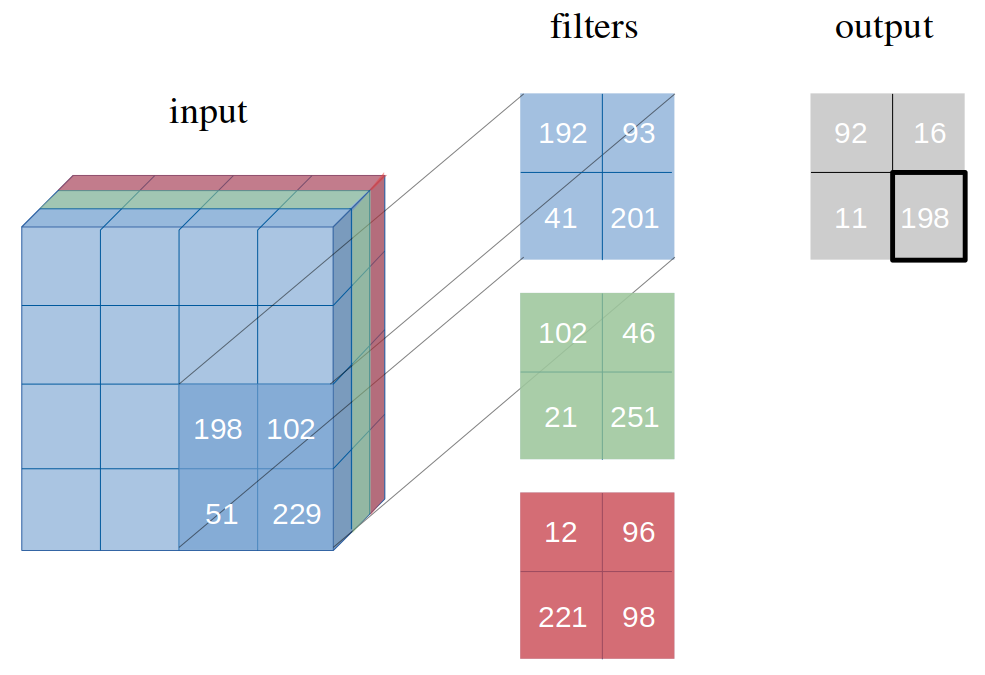}
\end{minipage}
\begin{minipage}{.4\textwidth}
\centering
\includegraphics[width=\linewidth]{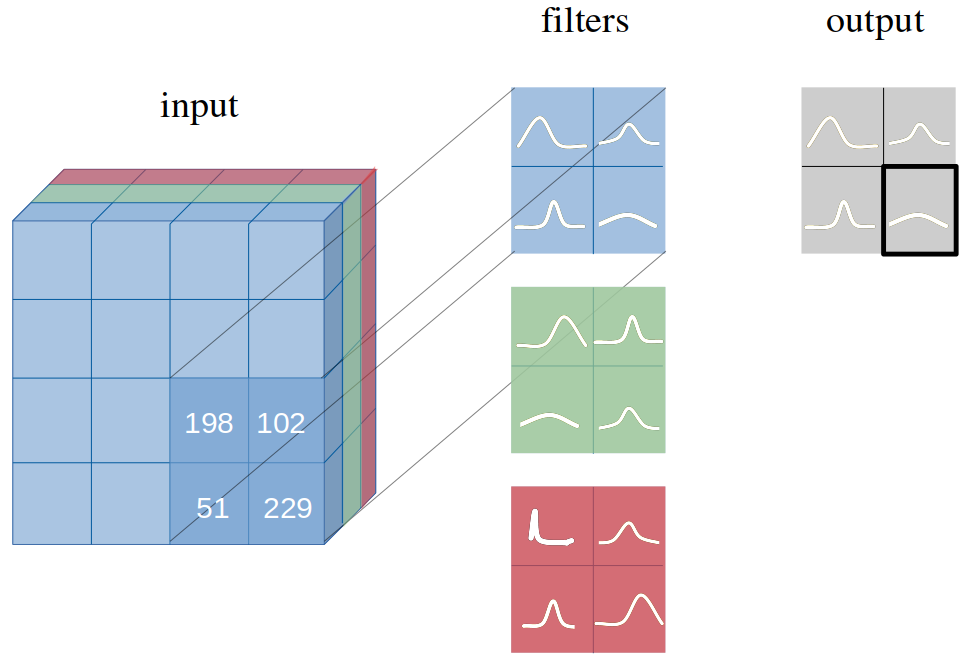}
\end{minipage}
\caption{Input image with exemplary pixel values, filters, and corresponding output with point-estimates (top) and probability distributions (bottom) over weights.\cite{shridhar2018bayesian}}
\label{fig:filter_scalar}
\end{figure} 
\begin{figure}[b!] 
\begin{center}
\includegraphics[width=\linewidth]{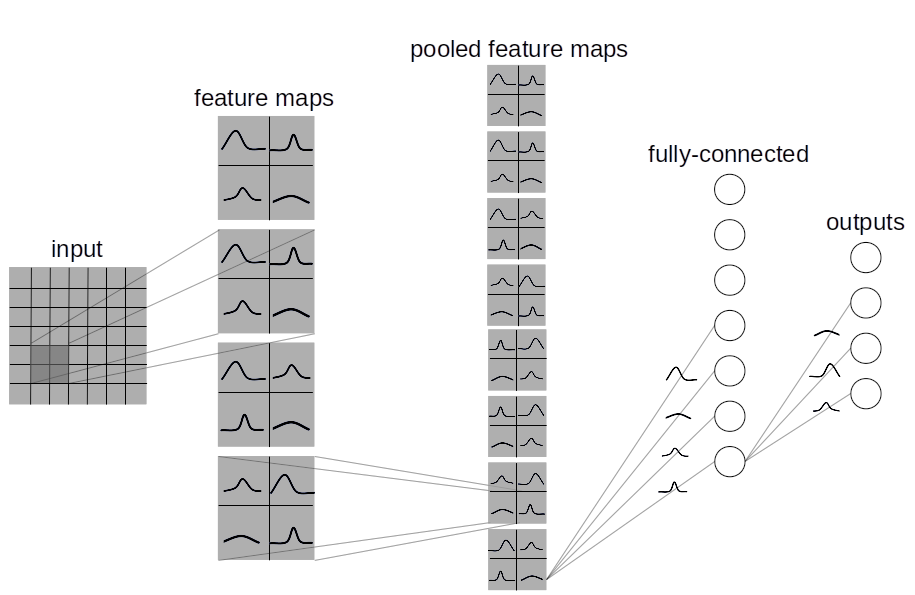}
\caption{Fully Bayesian perspective of an exemplary CNN. Weights in filters of convolutional layers, and weights in fully-connected layers have the form of a probability distribution. \cite{shridhar2018bayesian}}
\label{fig:CNNwithdist_grey}
\end{center}
\end{figure} 
\subsubsection{Local Reparameterization Trick for Convolutional Layers}
We utilise the local reparameterization trick \cite{kingma2015variational} and apply it to \acp{cnn}. Following \cite{kingma2015variational,neklyudov2018variance}, we do not sample the weights $w$, but we sample instead layer activations $b$ due to its consequent computational acceleration. The variational posterior probability distribution $q_{\theta}(w_{ijhw}|\mathcal{D})=\mathcal{N}(\mu_{ijhw},\alpha_{ijhw}\mu^2_{ijhw})$ (where $i$ and $j$ are the input, respectively output layers, $h$ and $w$ the height, respectively width of any given filter) allows to implement the local reparamerization trick in convolutional layers. This results in the subsequent equation for convolutional layer activations $b$:
\begin{equation}
    b_j=A_i\ast \mu_i+\epsilon_j\odot \sqrt{A^2_i\ast (\alpha_i\odot \mu^2_i)}
\end{equation}
where $\epsilon_j \sim \mathcal{N}(0,1)$, $A_i$ is the receptive field, $\ast$ signalises the convolutional operation, and $\odot$ the component-wise multiplication.

\subsubsection{Applying two Sequential Convolutional Operations (Mean and Variance)}
The crux of equipping a \ac{cnn} with probability distributions over weights instead of single point-estimates and being able to update the variational posterior probability distribution $q_{\theta}(w|\mathcal{D})$ by backpropagation lies in applying \textit{two} convolutional operations whereas filters with single point-estimates apply \textit{one}. As explained in the last chapter, we deploy the local reparametrization trick and sample from the output $b$. Since the output $b$ is a function of mean $\mu_{ijwh}$ and variance $\alpha_{ijhw}\mu^2_{ijhw}$ among others, we are then able to compute the two variables determining a Gaussian probability distribution, namely mean $\mu_{ijhw}$ and variance $\alpha_{ijhw}\mu^2_{ijhw}$, separately. 
\newline We do this in two convolutional operations: in the first, we treat the output $b$ as an output of a \ac{cnn} updated by frequentist inference. We optimize with Adam \cite{kingma2014adam} towards a single point-estimate which makes the validation accuracy of classifications increasing. We interpret this single point-estimate as the mean $\mu_{ijwh}$ of the variational posterior probability distributions $q_{\theta}(w|\mathcal{D})$. In the second convolutional operation, we learn the variance $\alpha_{ijhw}\mu^2_{ijhw}$. As this formulation of the variance includes the mean $\mu_{ijwh}$, only $\alpha_{ijhw}$ needs to be learned in the second convolutional operation \cite{molchanov2017variational}. In this way, we ensure that only one parameter is updated per convolutional operation, exactly how it would have been with a \ac{cnn} updated by frequentist inference. 
\newline In other words, while we learn in the first convolutional operation the MAP of the variational posterior probability distribution $q_{\theta}(w|\mathcal{D})$, we observe in the second convolutional operation how much values for weights $w$ deviate from this MAP. This procedure is repeated in the fully-connected layers. In addition, to accelerate computation, to ensure a positive non-zero variance $\alpha_{ijhw}\mu^2_{ijhw}$, and to enhance accuracy, we learn $\log \alpha_{ijhw}$ and use the \textit{Softplus} activation function as further described in the Emperical Analysis section.
\subsection{Uncertainty Estimation in CNN}
In classification tasks, we are interested in the predictive distribution $p_{\mathcal{D}}(y^*|x^*)$, where $x^*$ is an unseen data example and $y^*$ its predicted class. For a Bayesian neural network, this quantity is given by:
\begin{align}
p_{ \mathcal{D}}(y^*|x^*) = \int p_{w}(y^*|x^*) \ p_{\mathcal{D}}(w) \ dw
\end{align}
In \textit{Bayes by Backprop}, Gaussian distributions $q_{\theta}(w|\mathcal{D}) \sim \mathcal{N}(w|\mu, \sigma^2)$, where $\theta = \{ \mu, \sigma \}$ are learned with some dataset $\mathcal{D} = \{ x_{i}, y_{i} \}_{i=1}^{n}$ as we explained previously. Due to the discrete and finite nature of most classification tasks, the predictive distribution is commonly assumed to be a categorical. Incorporating this aspect into the predictive distribution gives us
\begin{align}
p_{\mathcal{D}}(y^*|x^*)& = \int \text{Cat}(y^*|f_w(x^*)) \mathcal{N}(w|\mu, \sigma^2) \ dw\\
&=  \int \prod_{c=1}^{C} f(x_{c}^*|w)^{y_{c}^*} \frac{1}{\sqrt{2\pi \sigma^2}} e^{-\frac{(w - \mu)^2}{2\sigma^2}} \ dw 
\end{align}
where $C$ is the total number of classes and $\sum_c f(x_{c}^*|w) = 1$.
\newline As there is no closed-form solution due to the lack of conjugacy between categorical and Gaussian distributions, we cannot recover this distribution. However, we can construct an unbiased estimator of the expectation by sampling from $q_{\theta}(w|\mathcal{D})$:
\begin{align}
\mathbb{E}_{q}[p_{\mathcal{D}}(y^*|x^*)] &= \int q_{\theta}(w|\mathcal{D}) \ p_w(y|x) \ dw \\ & \approx \frac{1}{T}\sum_{t=1}^{T} p_{w_t}(y^*|x^*)
\end{align}
where $T$ is the pre-defined number of samples.
This estimator allows us to evaluate the uncertainty of our predictions by the definition of variance, hence called \textit{predictive variance} and denoted as $\text{Var}_q$:
\begin{align} \label{variance}
\text{Var}_q\big( p(y^*|x^*) \big) = \mathbb{E}_q[yy^T] - \mathbb{E}_q[y]\mathbb{E}_q[y]^T
\end{align}
This quantity can be decomposed into the aleatoric and epistemic uncertainty \cite{kendall2017uncertainties,kwon2018uncertainty}. 
\begin{equation}
    \begin{aligned}
    \text{Var}_q\big( p(y^*|x^*) \big) &= \underbrace{\frac{1}{T} \sum_{t=1}^T \text{diag}(\hat{p}_t)-\hat{p}_t \ \hat{p}_t^T}_\text{aleatoric} \\ &+ \underbrace{\frac{1}{T}\sum_{t=1}^T (\hat{p}_t - \Bar{p}) (\hat{p}_t - \Bar{p})^T}_\text{epistemic}
    \end{aligned}
\end{equation}
where $\Bar{p} = \frac{1}{T}\sum_{t=1}^T \hat{p}_t$ and $\hat{p}_t = \text{Softmax}\big ( f_{w_{t}}(x^*) \big )$.

\begin{figure}[H] 
\centering
\begin{minipage}{.4\textwidth}
\centering
\includegraphics[width=\linewidth]{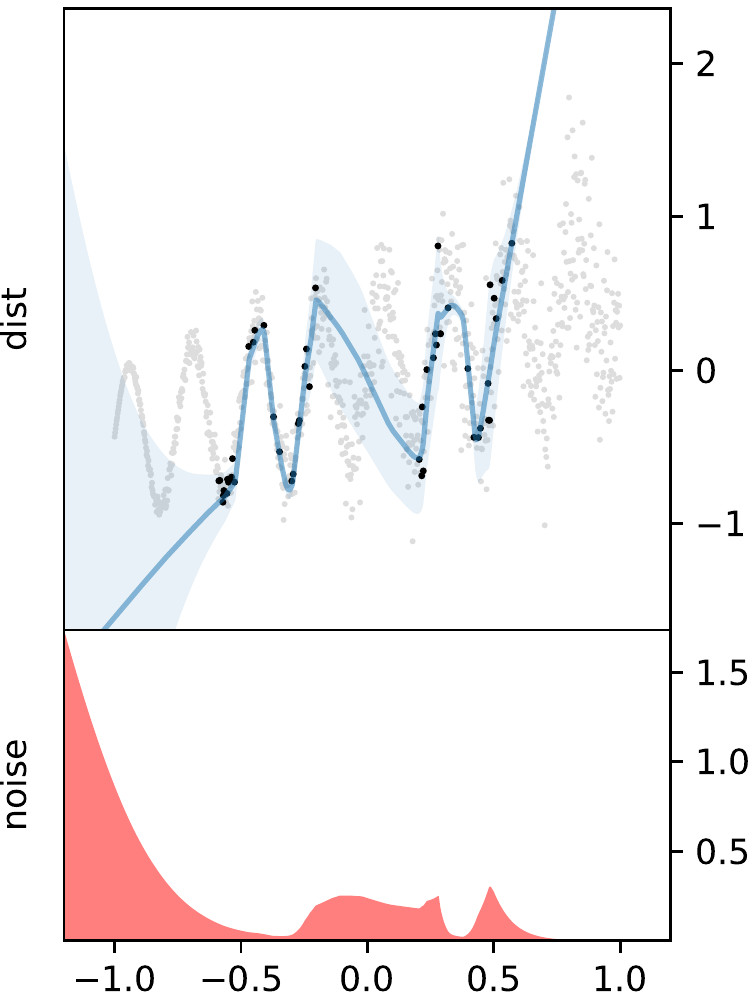}
\end{minipage}
\begin{minipage}{.4\textwidth}
\centering
\includegraphics[width=\linewidth]{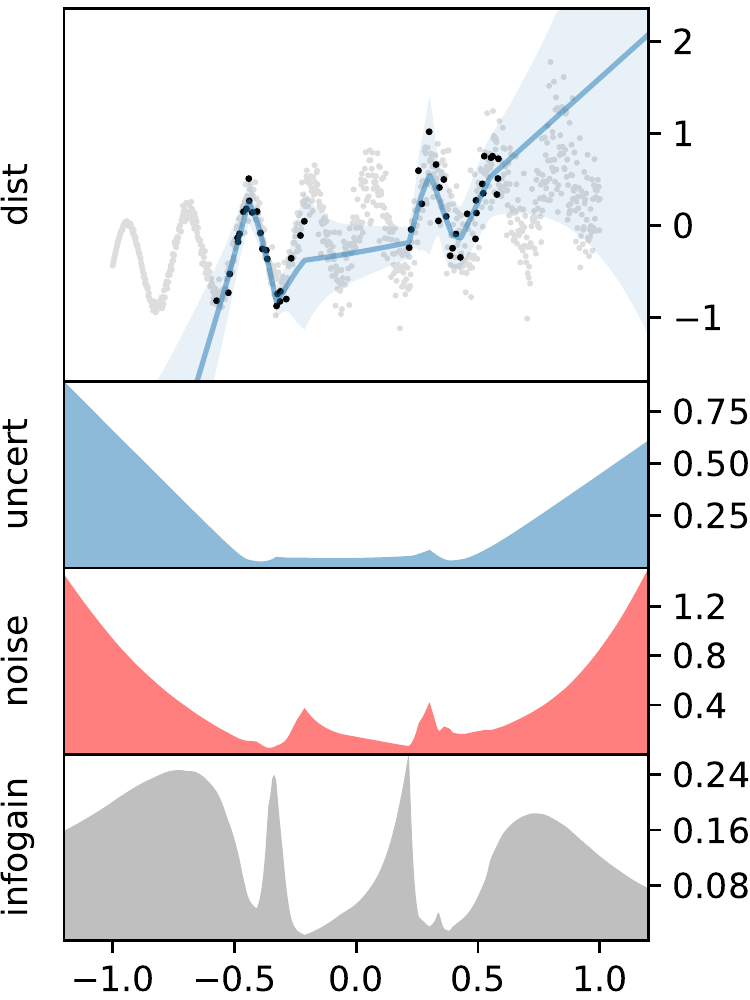}
\end{minipage}
\caption{Predictive distributions is estimated for a low-dimensional active learning task. The predictive distributions are visualized as mean and two standard deviations shaded. \textcolor{blue}{$\blacksquare$} shows the epistemic uncertainty and \textcolor{red}{$\blacksquare$} shows the aleatoric noise. Data points are shown in \textcolor{gray}{$\blacksquare$}.
\textbf{(Left)} A deterministic network conflates uncertainty as part of the noise and is overconfident outside of the data distribution.
\textbf{(Right)} A variational Bayesian neural network with standard normal prior represents uncertainty and noise separately but is overconfident outside of the training distribution as defined by \cite{hafner2018reliable}}
\label{fig:filter_scalar}
\end{figure} 

It is of paramount importance that uncertainty is split into aleatoric and epistemic quantities since it allows the modeller to evaluate the room for improvements: while aleatoric uncertainty (also known as statistical uncertainty) is merely a measure for the variation of ("noisy") data, epistemic uncertainty is caused by the model. Hence, a modeller can see whether the quality of the data is low (i.e. high aleatoric uncertainty), or the model itself is the cause for poor performances (i.e. high epistemic uncertainty). The former cannot be improved by gathering more data, whereas the latter can be done so. \cite{Kiureghian} \cite{kendall2017uncertainties}.

\subsection{Model Pruning}

Model pruning means the reduction in the model weights parameters to reduce the model overall non-zero weights, inference time and computation cost. In our work, a Bayesian Convolutional Network learns two weights, i.e: the mean and the variance compared to point estimate learning one single weight. This makes the overall number of parameters of a Bayesian Network twice as compared to the parameters of a point estimate similar architecture.

To make the Bayesian \acp{cnn} parameters equivalent to point estimate architecture, the number of filters in the Bayesian architectures is reduced to half. This makes up for the double learned parameters (mean and variance) against one in point estimates and makes the overall parameters equal for both networks. 

Another technique used is the usage of the \textit{L1 normalization} on the learned weights of every layer.  By L1 norm, we make the vector of learned weight in various model layers very sparse, as most of its components become close to zero, and at the same time, the remaining non-zero components capture the most important features of the data. We put a threshold and make the weights to be zero if the value falls below the threshold. We only keep the non zero weights and this way the model number of parameters is reduced without affecting the overall performance of the model. 

\section{Empirical Analysis}

\subsection{Experimentation Methodology} \label{experiments}

\subsubsection{Activation Function}

The originally chosen activation functions in all architectures are \textit{ReLU}, but we must introduce another, called \textit{Softplus}, see \eqref{softplus}, because of our method to apply two convolutional or fully-connected operations. As aforementioned, one of these is determining the mean $\mu$, and the other the variance $\alpha \mu^2$. Specifically, we apply the \textit{Softplus} function because we want to ensure that the variance $\alpha \mu^2$ never becomes zero. This would be equivalent to merely calculating the MAP, which can be interpreted as equivalent to a maximum likelihood estimation (MLE), which is further equivalent to utilising single point-estimates, hence frequentist inference. The \textit{Softplus} activation function is a smooth approximation of \textit{ReLU}. Although it is practically not influential, it has the subtle and analytically important advantage that it never becomes zero for $x \rightarrow -\infty$, whereas \textit{ReLU} becomes zero for $x \rightarrow -\infty$.
\\ 
\begin{equation}\label{softplus}
     \text{Softplus}(x) = \frac{1}{\beta} \cdot \log \big ( 1 + \exp(\beta \cdot x) \big )
\end{equation}
\\
where $\beta$ is by default set to $1$.
\newline All experiments are performed with the same hyper-parameters settings as stated in the Appendix.

\subsubsection{Network Architecture}

For all conducted experiments, we implement the foregoing description of Bayesian \acp{cnn} with variational inference in LeNet-5 \cite{lecun1998gradient} and AlexNet \cite{krizhevsky2012imagenet}. The exact architecture specifications can be found in the Appendix and in our GitHub repository\footnote{\url{https://github.com/kumar-shridhar/PyTorch-BayesianCNN}}.

\subsubsection{Objective Function}
To learn the objective function, we use \textit{Bayes by Backprop} \cite{graves2011practical, blundell2015weight}, which is a variational inference method to learn the posterior distribution on the weights $w \sim q_{\theta}(w|\mathcal{D})$ of a neural network from which weights $w$ can be sampled in backpropagation. 
It regularises the weights by minimising a compression cost, known as the variational free energy or the expected lower bound on the marginal likelihood.

 We tackled the problem of intractability in Chapter 2 and consequently, we arrive at the tractable cost function \eqref{cost} which is aimed to be optimized, i.e. minimised w.r.t. $\theta$, during training:
\begin{equation} \label{cost}
    \mathcal{F}(\mathcal{D}, \theta)\approx \sum_{i=1}^n \log q_{\theta}(w^{(i)}|\mathcal{D})-\log p(w^{(i)})-\log p(\mathcal{D}|w^{(i)})
\end{equation}
where $n$ is the number of draws.

Let's break the Objective Function \eqref{cost} and discuss in more details. 
\subsubsection{Variational Posterior }

The first term in the equation \eqref{cost} is the variational posterior. The variational posterior is taken as Gaussian distribution centred around mean $\mu$ and variance as $\sigma^2$. 

\begin{equation}
    q_{\theta}(w^{(i)}|\mathcal{D})= \prod_{i} \mathcal{N}(w_{i} | \mu,\sigma^2)
\end{equation}

We will take the log and the log posterior is defined as :

\begin{equation}
    log(q_{\theta}(w^{(i)}|\mathcal{D}))= \sum_{i}log \mathcal{N}(w_{i} | \mu,\sigma^2)
\end{equation}

\subsubsection{Prior}

The second term in the equation \eqref{cost} is the prior over the weights and we define the prior over the weights as a product of individual Gaussians :

\begin{equation}
    p(w^{(i)})= \prod_{i} \mathcal{N}(w_{i} | 0,\sigma_{p}^2)
\end{equation}

We will take the log and the log prior is defined as:

\begin{equation}
    log (p(w^{(i)}))= \sum_{i} log \mathcal{N}(w_{i} | 0,\sigma_{p}^2)
\end{equation}

\subsubsection{ Likelihood }

The final term of the equation \eqref{cost} $\log p(\mathcal{D}|w^{(i)})$ is the likelihood term and is computed using the softmax function.

\subsubsection{Parameter Initialization}

We use a Gaussian distribution and we store mean and variance values instead of just one weight. The way mean $\mu$ and variance $\sigma$ is computed is defined in the previous chapter. Variance cannot be negative and it is ensured by using \textit{softplus} as the activation function. We express variance $\sigma$ as $\sigma_{i}=softplus(\rho_{i})$ where $\rho$ is an unconstrained parameter. \\

We take the Gaussian distribution and initialize mean $\mu$ as 0 and variance $\sigma$ (and hence $\rho$) randomly. We observed mean centred around 0 and a variance starting with a big number and gradually decreasing over time. A good initialization can also be to put a restriction on variance and initialize it small. However, it might be data dependent and a good method for variance initialization is still to be discovered. We perform gradient descent over $\theta$ = ($\mu$, $\rho$), and individual weight $w_{i} \sim \mathcal{N} (w_{i} | \mu_{i}, \sigma_{i}$).  

\subsubsection{Optimizer}

For all our tasks, we take Adam optimizer \cite{kingma2014adam} to optimize the parameters. We also perform the local reparameterization trick as mentioned in the previous section and take the gradient of the combined loss function with respect to the variational parameters ($\mu$, $\rho$).

\subsubsection{Model Pruning}

We take the weights of all the layers of the network, apply an L1 norm over it and for all the weights value as zero or below a defined threshold are removed and the model is pruned. \\ Also, since the Bayesian \acp{cnn} has twice the number of parameters ($\mu$, $\sigma$) compared to a frequentist network (only 1 weight), we reduce the size of our network to half (AlexNet and LeNet- 5) by reducing the number of filters to half. The architecture used is mentioned in the Appendix.

\textit{Please note that it can be argued that reducing the number of filters to be half is a method for pruning or not. It can be seen as a method that reduces the number of overall parameters and hence can be thought of a pruning method in some sense. However, it is a subject to argument.} 

\subsection{Case Study 1: Small Datasets (MNIST, CIFAR-10)}

We train the networks with the MNIST dataset of handwritten digits \cite{lecun1998gradient}, and CIFAR-10 dataset \cite{krizhevsky2009learning} since these datasets serve widely as benchmarks for \acp{cnn}' performances. 

\subsubsection{MNIST}
The MNIST database \cite{lecun-mnisthandwrittendigit-2010} of handwritten digits have a training set of 60,000 examples and a test set of 10,000 examples. It is a subset of a larger set available from NIST. The digits have been size-normalized and centred in a fixed-size image of 28 by 28 pixels. Each image is grayscaled and is labelled with its corresponding class that ranges from zero to nine.
\newline

\subsubsection{CIFAR-10}
The CIFAR-10 are labelled subsets of the 80 million tiny images dataset \cite{Torralba:2008:MTI:1444381.1444403}. The CIFAR-10 dataset has a training dataset of 50,000 colour images in 10 classes, with 5,000 training images per class, each image 32 by 32 pixels large. There are 10000 images for testing. 
\newline

\subsection{Results}
First, we evaluate the performance of our proposed method, Bayesian \acp{cnn} with variational inference. Table \ref{tab:results} shows a comparison of validation accuracies (in percentage) for architectures trained by two disparate Bayesian approaches, namely variational inference, i.e. \textit{Bayes by Backprop} and Dropout as proposed by Gal and Ghahramani \cite{gal2015bayesian}.\\

We compare the results of these two approaches to frequentist inference approach for both the datasets. Bayesian \acp{cnn} trained by variational inference achieve validation accuracies comparable to their counter-architectures trained by frequentist inference. On MNIST, validation accuracies of the two disparate Bayesian approaches are comparable, but a Bayesian LeNet-5 with Dropout achieves a considerable higher validation accuracy on CIFAR-10, although we were not able to reproduce these reported results.
\begin{table}[H]
\tiny
    \centering
    \renewcommand{\arraystretch}{1.5}
    \resizebox{\linewidth}{!}{
    \begin{tabular}{ l  c  c  c  c } 
     \hline
      \empty & MNIST & CIFAR-10   \\ [0.75ex]
     \hline
     Bayesian AlexNet (with VI) & 99 & 73   \\
     
     Frequentist AlexNet & 99 & 73   \\
     \hdashline
     Bayesian LeNet-5 (with VI) & 98 & 69   \\
     
     Frequentist LeNet-5 & 98 & 68   \\
     \hdashline
     Bayesian LeNet-5 (with Dropout) & 99.5 & 83  \\ 
     \hline \\
    \end{tabular}}
    \renewcommand{\arraystretch}{1.5}
    \caption{Comparison of validation accuracies (in percentage) for different architectures with variational inference (VI), frequentist inference and Dropout as a Bayesian approximation as proposed by Gal and Ghahramani \cite{gal2015bayesian} for MNIST, and CIFAR-10.}
    \label{tab:results}
\end{table}

\begin{figure}[t!] 
\begin{center}
\includegraphics[width=\linewidth]{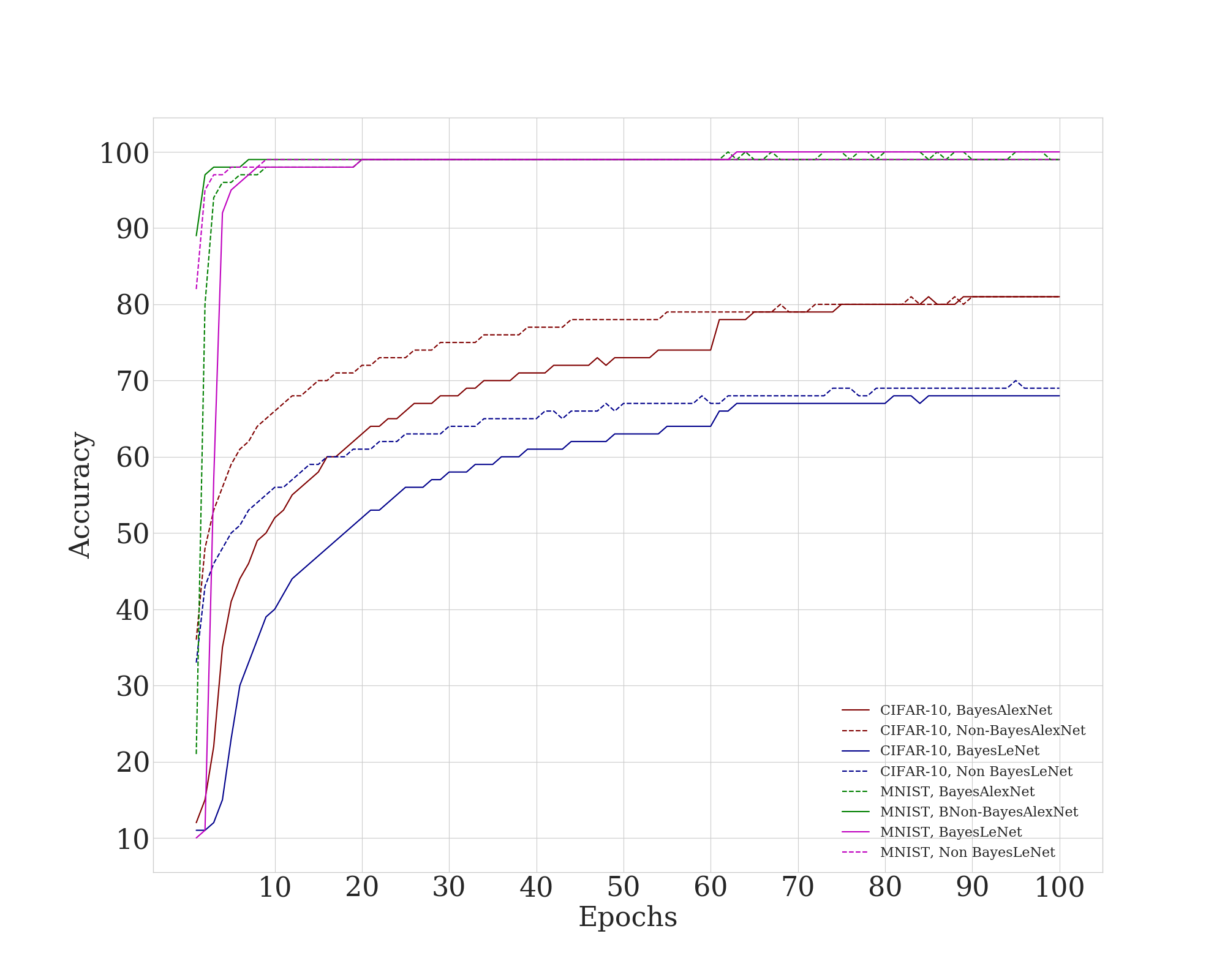}
\caption{Comparison of Validation Accuracies of Bayesian AlexNet and LeNet-5 with frequentist approach on MNIST and CIFAR-10 datasets}
\label{fig:MnistCIFAR10reesults}
\end{center}
\end{figure} 

Figure \ref{fig:MnistCIFAR10reesults} shows the validation accuracies of Bayesian vs Non-Bayesian \acp{cnn}. One thing to observe is that in initial epochs, Bayesian \acp{cnn} trained by variational inference start with a low validation accuracy compared to architectures trained by frequentist inference. This must deduce from the initialization of the variational posterior probability distributions $q_{\theta}(w|\mathcal{D})$ as uniform distributions, while initial point-estimates in architectures trained by frequentist inference are randomly drawn from a standard Gaussian distribution. (For uniformity, we changed the initialization of frequentist architectures from Xavier initialization to standard Gaussian). The latter initialization method ensures the initialized weights are neither too small nor too large. In other words, the motivation of the latter initialization is to start with weights such that the activation functions do not let them begin in saturated or dead regions. This is not true in case of uniform distributions and hence, Bayesian \acp{cnn}' starting validation accuracies can be comparably low.

\begin{figure}[H] 
\begin{center}
\includegraphics[width=\linewidth]{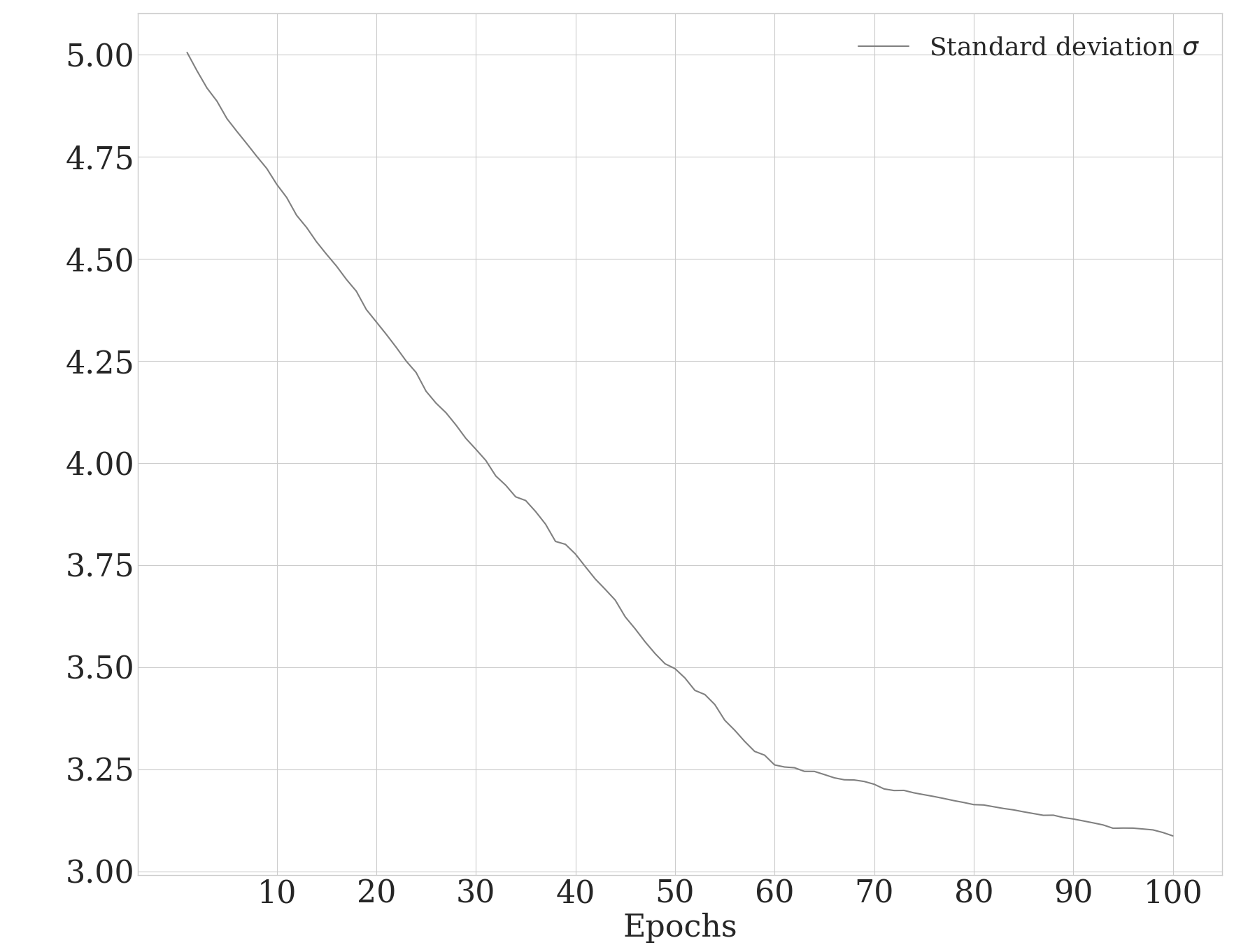}
\caption{Convergence of the standard deviation of the Gaussian variational posterior probability distribution $q_{\theta}(w|\mathcal{D})$ of a random model parameter at epochs 1, 5, 20, 50, and 100. MNIST is trained on Bayesian LeNet-5.}
\label{fig:std_CNN}
\end{center}
\end{figure}

\begin{figure}[H] 
\begin{center}
\includegraphics[width=\linewidth]{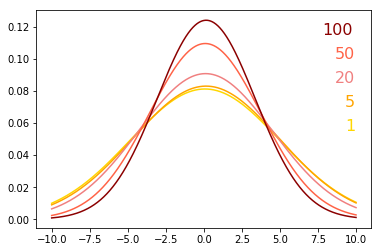}
\caption{Convergence of the Gaussian variational posterior probability distribution $q_{\theta}(w|\mathcal{D})$ of a random model parameter at epochs 1, 5, 20, 50, and 100. CIFAR-10 is trained on Bayesian LeNet-5.}
\label{fig:distribution}
\end{center}
\end{figure}

Figure \ref{fig:std_CNN} displays the convergence of the standard deviation $\sigma$ of the variational posterior probability distribution $q_{\theta}(w|\mathcal{D})$ of a random model parameter over epochs. As aforementioned, all prior probability distributions $p(w)$ are initialized as uniform distributions. The variational posterior probability distributions $q_{\theta}(w|\mathcal{D})$ are approximated as Gaussian distributions which become more confident as more data is processed - observable by the decreasing standard deviation over epochs in Figure \ref{fig:std_CNN}. Although the validation accuracy for MNIST on Bayesian LeNet-5 has already reached 99\%, we can still see a fairly steep decrease in the parameter's standard deviation. In Figure \ref{fig:distribution}, we plot the actual Gaussian variational posterior probability distributions $q_{\theta}(w|\mathcal{D})$ of a random parameter of LeNet-5 trained on CIFAR-10 at some epochs.

Figure \ref{fig:distribution} displays the convergence of the Gaussian variational probability distribution of a weight taken randomly from the first layer of LeNet-5 architecture. The architecture is trained on CIFAR-10 dataset with uniform initialization.

\subsection{Case Study 2: Large Dataset (CIFAR-100)}

\subsubsection{CIFAR-100}
This dataset is similar to the CIFAR-10 and is a labelled subset of the 80 million tiny images dataset \cite{Torralba:2008:MTI:1444381.1444403}. The dataset has 100 classes containing 600 images each. There are 500 training images and 100 validation images per class. The images are coloured with a resolution of 32 by 32 pixels.

\subsection{Results}

\begin{figure}[H] 
\begin{center}
\includegraphics[width=\linewidth]{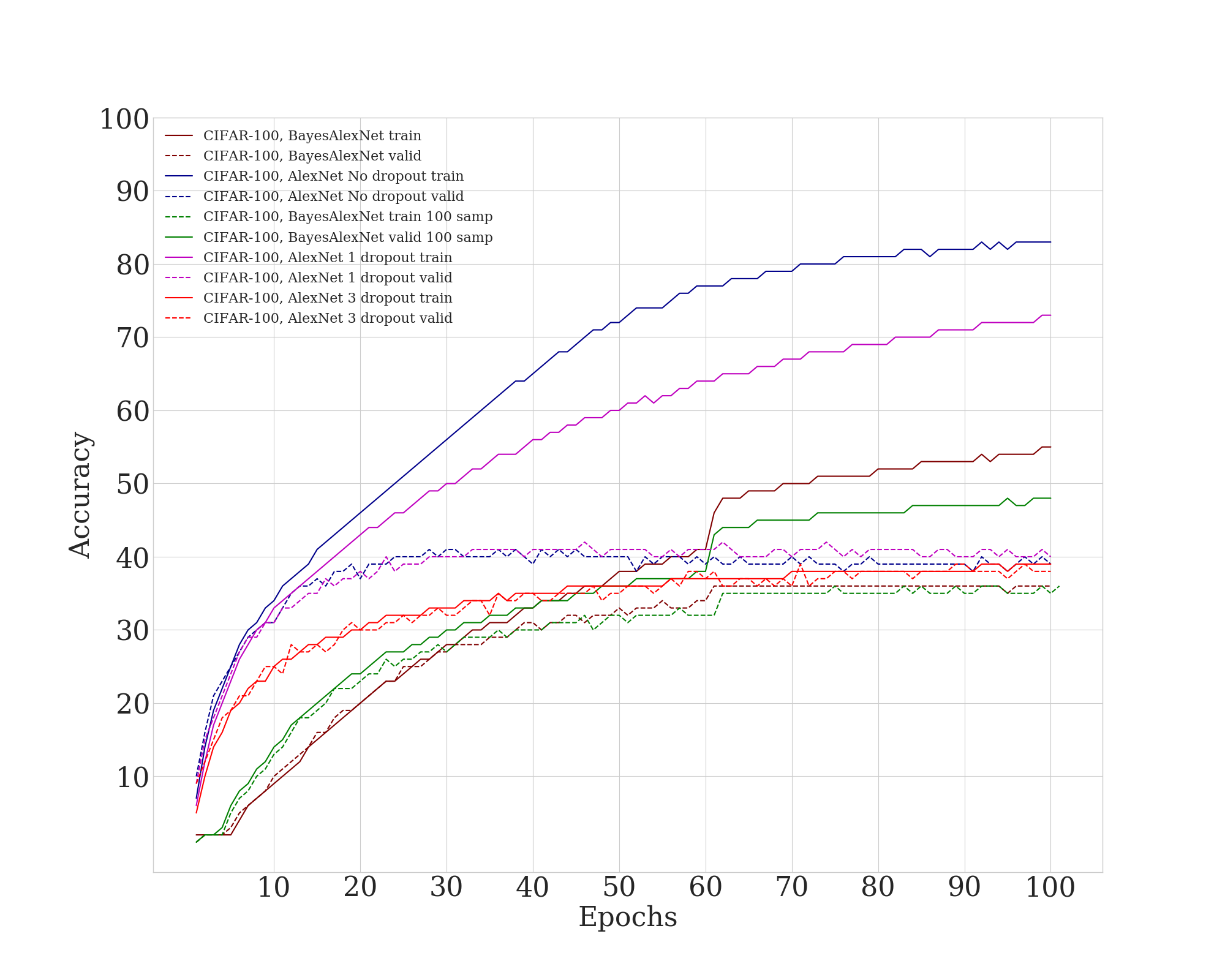}
\caption{Comparison of Training and Validation Accuracies of Bayesian AlexNet and LeNet-5 with frequentist approach with and without dropouts on CIFAR-100 datasets}
\label{fig:regularization}
\end{center}
\end{figure}

In Figure \ref{fig:regularization}, we show how Bayesian networks incorporate naturally effects of regularization, exemplified on AlexNet. While an AlexNet trained by frequentist inference without any regularization overfits greatly on CIFAR-100, an AlexNet trained by Bayesian inference on CIFAR-100 does not. This is evident from the high value of training accuracy for frequentist approach with no dropout or 1 layer dropout. Bayesian CNN performs equivalently to an AlexNet trained by frequentist inference with three layers of Dropout after the first, fourth, and sixth layers in the architecture.
Another thing to note here is that the Bayesian CNN with 100 samples overfits slightly lesser compared to Bayesian CNN with 25 samples. However, a higher sampling number on a smaller dataset didn't prove useful and we stuck with 25 as the number of samples for all other experiments.

\begin{table}[H]
\tiny
    \centering
    \renewcommand{\arraystretch}{1.5}
    \resizebox{\linewidth}{!}{
    \begin{tabular}{ l  c  c  } 
     \hline
      \empty & CIFAR-100  \\ [0.75ex]
     \hline
     Bayesian AlexNet (with VI)  & 36  \\
     
     Frequentist AlexNet & 38  \\

     Bayesian LeNet-5 (with VI) &  31  \\
     
     Frequentist LeNet-5  & 33  \\
     \hline \\
    \end{tabular}}
    \renewcommand{\arraystretch}{1.5}
    \caption{Comparison of validation accuracies (in percentage) for different architectures with variational inference (VI), frequentist inference and Dropout as a Bayesian approximation as proposed by Gal and Ghahramani \cite{gal2015bayesian} for MNIST, CIFAR-10, and CIFAR-100.}
    \label{tab:resultsCIFAR-100}
\end{table}

Table \ref{tab:tableCIFAR100} shows a comparison of the training and validation accuracies for AlexNet with Bayesian approach and frequentist approach. The low gap between the training and validation accuracies shows the robustness of Bayesian approach towards overfitting and shows how Bayesian approach without being regularized overfits lesser as compared to frequentist architecture with no or one dropout layer. The results are comparable with AlexNet architecture with 3 dropout layers.

\begin{table}[H]
\tiny
    \centering
    \renewcommand{\arraystretch}{1.5}
    \resizebox{\linewidth}{!}{
    \begin{tabular}{ l  c  c  c  c } 
     \hline
      \empty & Training Accuracy & Validation Accuracy   \\ [0.75ex]
     \hline
     Frequentist AlexNet (No dropout) & 83 & 38   \\
     
     Frequentist AlexNet (1 dropout layer) & 72 & 40   \\
     
      Frequentist AlexNet (3 dropout layer) & 39 & 38   \\
     \hdashline
     Bayesian AlexNet (25 num of samples) & 54 & 37   \\
     
     Bayesian AlexNet (100 num of samples) & 48 & 37   \\
     \hline \\
    \end{tabular}}
    \renewcommand{\arraystretch}{1.5}
    \caption{Comparison of training and validation accuracies (in percentage) for AlexNet architecture with variational inference (VI) and frequentist inference for CIFAR-100.}
    \label{tab:tableCIFAR100}
\end{table}

\subsection{Uncertainity Estimation}

Finally, Table \ref{tab:uncertainty} compares the means of aleatoric and epistemic uncertainties for a Bayesian LeNet-5 with variational inference on MNIST and CIFAR-10. The aleatoric uncertainty of CIFAR-10 is about twenty times as large as that of MNIST. Considering that the aleatoric uncertainty measures the irreducible variability and depends on the predicted values, a larger aleatoric uncertainty for CIFAR-10 can be directly deduced from its lower validation accuracy and may be further due to the smaller number of training examples. The epistemic uncertainty of CIFAR-10 is about fifteen times larger than that of MNIST, which we anticipated since epistemic uncertainty decreases proportionally to validation accuracy. 
\begin{table}[H]
\tiny
    \centering
    \renewcommand{\arraystretch}{1.5}
    \resizebox{\linewidth}{!}{
    \begin{tabular}{ l  c  c  c  } 
     \hline
      \empty & Aleatoric uncertainty &  Epistemic uncertainty  \\ [0.75ex]
     \hline
     Bayesian LeNet-5 (MNIST) & 0.0096 & 0.0026   \\
     
     Bayesian LeNet-5 (CIFAR-10) & 0.1920 & 0.0404   \\
     \hline \\
    \end{tabular}} 
    \renewcommand{\arraystretch}{1.5}
    \caption{Aleatoric and epistemic uncertainty for Bayesian LeNet-5 calculated for MNIST and CIFAR-10, computed as proposed by Kwon et al. \cite{kwon2018uncertainty}.}
    \label{tab:uncertainty}
\end{table}

\subsection{Model Pruning}

\subsubsection{Halving the Number of Filters}

For every parameter for a frequentist inference network, Bayesian \acp{cnn} has two parameters ($\mu$, $\sigma$). Halving the number of parameters of Bayesian AlexNet ensures the number of parameters of it is comparable with a frequentist inference network. The number of filters of ALexNet is halved and a new architecture called AlexNetHalf is defined in Figure 5.4. 

\begin{table}[h!]
    \centering
    \renewcommand{\arraystretch}{2}
    \begin{tabular}{c c c c c c} 
 \hline
 layer type & width & stride & padding & input shape & nonlinearity \\ [0.5ex] 
 \hline
 convolution ($11\times11$) & 32 & 4 & 5 & $M\times3\times32\times32$ & Softplus \\ 
 
 max-pooling ($2\times2$) & \empty & 2 & 0 & $M\times32\times32\times32$ & \empty \\
 
 convolution ($5\times5$) & 96 & 1 & 2 & $M\times32\times15\times15$ & Softplus \\
 
 max-pooling ($2\times2$) & \empty & 2 & 0 & $M\times96\times15\times15$ & \empty \\
 
 convolution ($3\times3$) & 192 & 1 & 1 & $M\times96\times7\times7$ & Softplus \\
 
 convolution ($3\times3$) & 128 & 1 & 1 & $M\times192\times7\times7$ & Softplus \\
 
 convolution ($3\times3$) & 64 & 1 & 1 & $M\times128\times7\times7$ & Softplus \\
 
 max-pooling ($2\times2$) & \empty & 2 & 0 & $M\times64\times7\times7$ & \empty \\
 
 fully-connected & 64 & \empty & \empty & $M\times64$ & \empty \\ [1ex] 
 \hline
\end{tabular}
\renewcommand{\arraystretch}{1.5}
\label{tab:AlexNetHalfArchitecture}
\caption{AlexNetHalf with number of filters halved compared to the original architecture.}
\end{table}

The AlexNetHalf architecture was trained and validated on the MNIST, CIFAR10 and CIFAR100 dataset and the results are shown in Table \ref{tab:resultsAlexNetHalf}. The accuracy of pruned AlexNet with only half the number of filters compared to the normal architecture shows an accuracy gain of 6 per cent in case of CIFAR10 and equivalent performance for MNIST and CIFAR100 datasets. A lesser number of filters learn the most important features which proved better at inter-class classification could be one of the explanations for the rise in accuracy. However, upon visualization of the filters, no distinct clarification can be made to prove the previous statement. \\ 
Another possible explanation could be the model is generalizing better after the reduction in the number of filters ensuring the model is not overfitting and validation accuracy is comparatively higher. CIFAR-100 higher validation accuracy on ALexNetHalf and a lower training accuracy than Bayesian AlexNet proves the theory. Using a lesser number of filters further enhances the regularization effect and makes the model more robust against overfitting. Similar results have been achieved by Narang \cite{DBLP:journals/corr/NarangDSE17} in his work where a pruned model achieved better accuracy compared to the original architecture in a speech recognition task. Suppressing or removing the weights that have lesser or no contribution to the prediction makes the model rely its prediction on the most prominent and unique features and hence improves the prediction accuracy.

\begin{table}[H]
\tiny
    \centering
    \renewcommand{\arraystretch}{1.5}
    \resizebox{\linewidth}{!}{
    \begin{tabular}{ l  c  c  c  c } 
     \hline
      \empty & MNIST & CIFAR-10 & CIFAR-100 \\ [0.75ex]
     \hline
     Bayesian AlexNet (with VI) & 99 & 73 & 36 \\
     
     Frequentist AlexNet & 99 & 73 & 38  \\
     
     Bayesian AlexNetHalf (with VI) & 99 & 79 & 38 \\
     
     \hline \\
    \end{tabular}}
    \renewcommand{\arraystretch}{1.5}
    \caption{Comparison of validation accuracies (in percentage) for AlexNet with variational inference (VI), AlexNet with frequentist inference and AlexNet with half number of filters halved for MNIST, CIFAR-10 and CIFAR-100 datasets.}
    \label{tab:resultsAlexNetHalf}
\end{table}

\subsubsection{Applying L1 Norm}

L1 norm induces sparsity in the trained model parameters and sets some values to zero. We trained a model to some epochs (number of epochs differs across datasets as we applied early stopping when validation accuracy remains unchanged for 5 epochs). We removed the zero-valued parameters of the learned weights and keep the non-zero parameters for a trained Bayesian AlexNet on MNIST and CIFAR-10 datasets. We pruned the model to make the number of parameters in a Bayesian Network comparable to the number of parameters in the point-estimate architecture. \\ Table \ref{tab:resultsL1Norm} shows the comparison of validation accuracies of the applied L1 Norm AlexNet Bayesian architecture with Bayesian AlexNet architecture and with AlexNet frequentist architecture. We got comparable results on MNIST and CIFAR10 with the experiments and the results are shown in Table \ref{tab:resultsL1Norm}

\begin{table}[H]
\tiny
    \centering
    \renewcommand{\arraystretch}{1.5}
    \resizebox{\linewidth}{!}{
    \begin{tabular}{ l  c  c  c  } 
     \hline
      \empty & MNIST & CIFAR-10  \\ [0.75ex]
     \hline
     Bayesian AlexNet (with VI) & 99 & 73  \\
     
     Frequentist AlexNet & 99 & 73   \\
     
     Bayesian AlexNet with L1 Norm (with VI) & 99 & 71  \\
     
     \hline \\
    \end{tabular}}
    \renewcommand{\arraystretch}{1.5}
    \caption{Comparison of validation accuracies (in percentage) for AlexNet with variational inference (VI), AlexNet with frequentist inference and BayesianAlexNet with L1 norm applied for MNIST and CIFAR-10 datasets.}
    \label{tab:resultsL1Norm}
\end{table}

One thing to note here is that the numbers of parameters of Bayesian Network after applying L1 norm is not necessarily equal to the number of parameters in the frequentist AlexNet architecture. It depends on the data size and the number of classes. However, the number of parameters in the case of MNIST and CIFAR-10 are pretty comparable and there is not much reduction in the accuracy either. Also, the early stopping was applied when there is no change in the validation accuracy for 5 epochs and the model was saved and later pruned with the application of L1 norm.

\subsection{Training Time}

Training time of a Bayesian \acp{cnn} is twice of a frequentist network with similar architecture when the number of samples is equal to one. In general, the training time of a Bayesian \acp{cnn}, $T$ is defined as:
\begin{align}
T = 2 * number\_of\_samples * t
\end{align}
where $t$ is the training time of a frequentist network. 
The factor of 2 is present due to the double learnable parameters in a Bayesian CNN network i.e. mean and variance for every single point estimate weight in the frequentist network.

However, there is no difference in the inference time for both the networks. 

\section{Application}

\subsection{BayesCNN for Image Super Resolution}

The task referred to as Super Resolution (SR) is the recovery of a High-Resolution (HR) image from a given Low-Resolution (LR) image. It is applicable to many areas like medical imaging \cite{10.1007/978-3-642-40760-4_2}, face recognition \cite{1203152} and so on.

There are many ways to do a single image super-resolution and detailed benchmarks of the methods are provided by Yang \cite{Yang2014SingleImageSA}. Following are the major ways to do a single image super-resolution:\\ 

\textbf{Prediction Models}: These models generate High-Resolution images
from Low-Resolution inputs through a predefined mathematical formula. No training data is needed for such models. Interpolation-based methods (bilinear, bicubic, and Lanczos) generate HR pixel intensities by weighted averaging neighbouring LR pixel values are good examples of this method.\\

\textbf{Edge Based Methods}: Edges are one of the most important features for any computer vision task. The High-Resolution images learned from the edge features high-quality edges and good sharpness. However, these models lack good colour and texture information.\\

\textbf{Patch Based Methods}: Cropped patches from Low-Resolution images and High-Resolution images are taken from training dataset to learn some mapping function. The overlapped patches are averaged or some other techniques like Conditional Random Fields \cite{lafferty2001conditional} can be used for better mapping of the patches.

\subsubsection{Our Approach}

We build our work upon \cite{DBLP:journals/corr/ShiCHTABRW16} work that shows that performing Super Resolution work in High-Resolution space is not the optimal solution and it adds the computation complexity. We used a Bayesian Convolutional Neural Network to extract features in the Low-Resolution space. We use an efficient sub-pixel convolution layer, as proposed by \cite{DBLP:journals/corr/ShiCHTABRW16}, which learns an array of upscaling filters to upscale the final Low-Resolution feature maps into the High-Resolution output. This replaces the handcrafted bicubic filter in the Super Resolution pipeline with more complex upscaling filters specifically trained for each feature map, and also reduces the computational complexity of the overall Super Resolution operation.

The hyperparameters used in the experiments are mentioned in the Appendix A section in details.

\begin{figure*}[htbp]
\begin{center}
\includegraphics[width=1.0\linewidth]{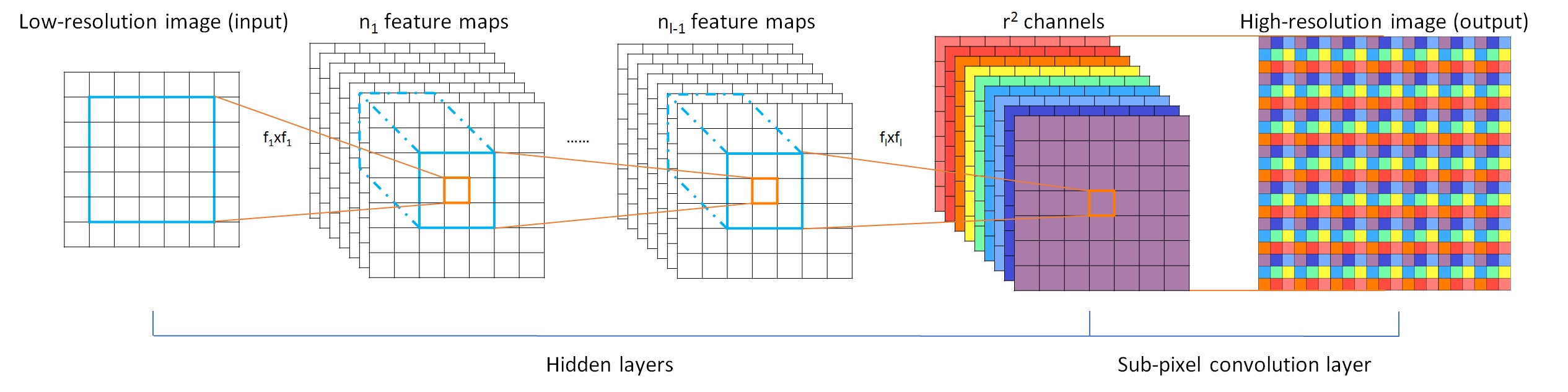}
\caption{The proposed efficient sub-pixel convolutional neural network (ESPCN) \cite{DBLP:journals/corr/ShiCHTABRW16}, with two convolution layers for feature maps extraction, and a sub-pixel convolution layer that aggregates the feature maps from Low Resolution space and builds the Super Resolution image in a single step.}
\label{fig:networkstructure}
\end{center}
\end{figure*}

We used a four-layer convolutional model as mentioned in the paper \cite{DBLP:journals/corr/ShiCHTABRW16}. We replaced the convolution layer by Bayesian convolution layer and changed the forward pass that now computes the mean, variance and KL divergence. The PixelShuffle layer is kept same as provided by PyTorch and no changes have been made there.  

\begin{table}[H]
    \centering
    \renewcommand{\arraystretch}{2}
    \begin{tabular}{c c c c } 
 \hline
 layer type & width & stride & padding  \\ [0.5ex] 
 \hline
 convolution ($5\times5$) & 64 & 1 & 2 \\
 
 convolution ($3\times3$) & 64 & 1 & 1 \\

 convolution ($3\times3$) & 32 & 1 & 1 \\
 
 convolution ($3\times3$) & upscale factor * * 2 & 1 & 1  \\ [1ex] 
 \hline
\end{tabular}
\renewcommand{\arraystretch}{1.5}
\label{tab:SuperResolutionArchitecture}
\caption{Network Architecture for Bayesian Super Resolution}
\end{table}

Where \textit{upscale factor} is defined as a parameter. For our experiments, we take upscale factor = 3. 

\subsubsection{Empirical Analysis}

The Network architecture was trained on BSD300 dataset \cite{MartinFTM01} provided by the Berkeley Computer Vision Department. The dataset is very popular for Image Super-Resolution task and thus the dataset is used to compare the results with other work in the domain. 

\begin{figure}[H]
\begin{center}
\includegraphics[height=.28\textheight]{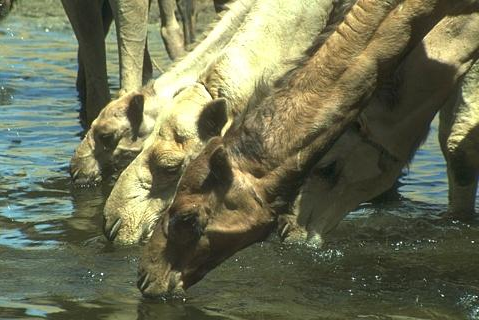}
\label{fig:CamelLR}
\caption{Sample image in Low Resolution image space taken randomly from BSD 300 \cite{MartinFTM01} dataset.}
\end{center}
\end{figure}

\begin{figure}[H]
\begin{center}
\includegraphics[height=.38\textheight]{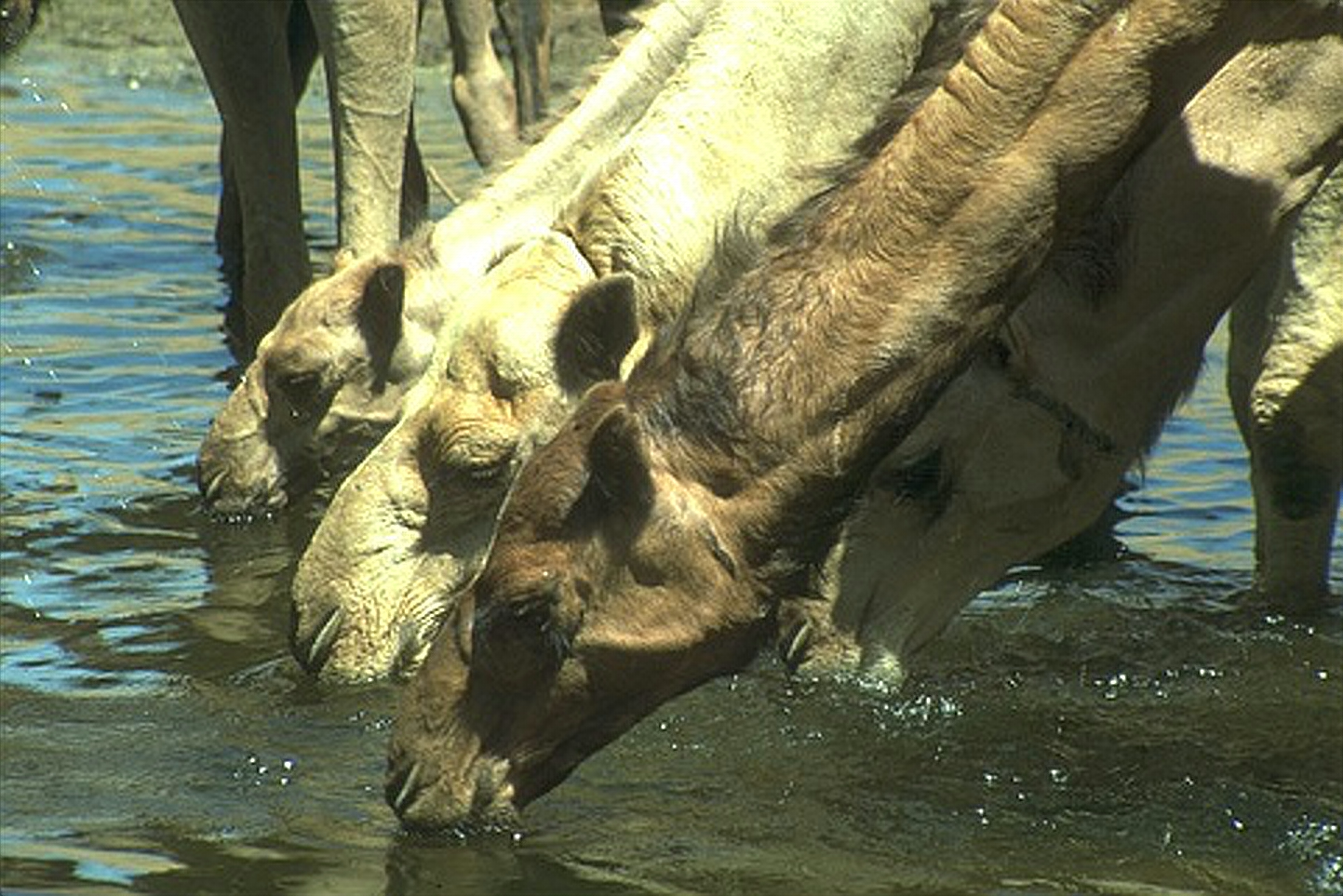}
\label{fig:CamelSR}
\caption{Generated Super Resolution Image scaled to 40 percent to fit }
\end{center}
\end{figure}

The generated results with Bayesian Network is compared with the original paper and the results are comparable in terms of the number and the quality of the image generated. This application was to prove the concept that the Bayesian Networks can be used for the task of Image Super Resolution. Furthermore, the results are pretty good. 

Some more research is needed in the future to achieve state-of-the-art results in this domain which is out of the scope of this thesis work.

\subsection{BayesCNN for Generative Adversarial Networks}

Generative Adversarial Networks (GANs) \cite{goodfellow2014generative} can be used for two major tasks: to learn good feature representations by using the generator and discriminator networks as feature extractors and to generate natural images. The learned feature representation or generated images can reduce the number of images substantially for a computer vision supervised task. However, GANs were quite unstable to train in the past and that is why we base our work on the stable GAN architecture namely Deep Convolutional GANs (DCGAN) \cite{DBLP:journals/corr/RadfordMC15}. We use the trained Bayesian discriminators for image classification tasks, showing competitive performance with the normal DCGAN architecture.

\subsubsection{Our Approach}

We based our work on the paper:  Unsupervised Representation Learning with Deep Convolutional Generative Adversarial Networks by  \cite{DBLP:journals/corr/RadfordMC15}. We used the architecture of a deep convolutional generative adversarial networks (DCGANs) that learns a hierarchy of representations from object parts to scenes in both the generator and discriminator.
The generator used in the Network is shown in Table \ref{tab:GeneratorArchitecture}. The architecture is kept similar to the architecture used in DCGAN paper \cite{DBLP:journals/corr/RadfordMC15}. Table \ref{tab:DiscriminatorArchitecture} shows the discriminator network with Bayesian Convolutional Layers. 

\begin{table}[H]
    \centering
    \renewcommand{\arraystretch}{2}
    \begin{tabular}{c c c c c} 
 \hline
 layer type & width & stride & padding & nonlinearity \\ [0.5ex] 
 \hline
 ConvolutionTranspose ($4\times4$) & ngf * 8 & 1 & 0  & ReLU \\

 ConvolutionTranspose ($4\times4$) & ngf * 4 & 2 & 1  & ReLU \\

 ConvolutionTranspose ($4\times4$) & ngf * 2 & 2 & 1 & ReLU \\
 
 ConvolutionTranspose ($4\times4$) & ngf & 2 & 1  & ReLU \\
 
 ConvolutionTranspose ($4\times4$) & nc & 2 & 1 & TanH \\ [1ex] 
 \hline
\end{tabular}
\renewcommand{\arraystretch}{1}
\caption{Generator architecture as defined in the paper. \cite{DBLP:journals/corr/RadfordMC15}}
\label{tab:GeneratorArchitecture}
\end{table}

where \textit{ngf} is the number of generator filters which is chosen to be 64 in our work and \textit{nc} is the number of output channels which is set to 3. 

\begin{table}[H]
    \centering
    \renewcommand{\arraystretch}{2}
    \begin{tabular}{c c c c c} 
 \hline
 layer type & width & stride & padding & nonlinearity \\ [0.5ex] 
 \hline
 Convolution ($4\times4$) & ndf & 2 & 1  & LeakyReLU \\

 Convolution($4\times4$) & ndf * 2 & 2 & 1  & LeakyReLU \\

 Convolution ($4\times4$) & ndf * 4 & 2 & 1 & LeakyReLU \\
 
 Convolution ($4\times4$) & ndf * 8 & 2 & 1  & leakyReLU \\
 
 ConvolutionTranspose ($4\times4$) & 1 & 1 & 0 & Sigmoid \\ [1ex] 
 \hline
\end{tabular}
\renewcommand{\arraystretch}{1}
\caption{Discriminator architecture with Bayesian Convolutional layers}
\label{tab:DiscriminatorArchitecture}
\end{table}

where \textit{ndf} is the number of discriminator filters and is set to 64 as default for all our experiments. 

\subsubsection{Empirical Analysis}

The images were taken directly and no pre-processing was applied to any of the images. Normalization was applied with value 0.5 to make the data mean centred. A batch size of 64 was used along with Adam \cite{kingma2014adam} as an optimizer to speed up the training. All weights were initialized from a zero-centred Normal distribution with standard deviation equal to 1. We also used LeakyReLU as mentioned in the original DCGAN paper \cite{DBLP:journals/corr/RadfordMC15}. The slope of the leak in LeakyReLU was set to 0.2 in all models. We used the learning rate of 0.0001, whereas in paper 0.0002 was used instead. Additionally, we found leaving the momentum term $\beta_1$ at the suggested value of 0.9 resulted in training oscillation and instability while reducing it to 0.5 helped stabilize training (also taken from original paper \cite{DBLP:journals/corr/RadfordMC15}).

The hyperparameters used in the experiments are mentioned in the Appendix A section in details.
The fake results of the generator after 100 epochs of training is shown in Figure 6.4. To compare the results, real samples are shown in Figure 6.5. The loss in case of a Bayesian network is higher as compared to the DCGAN architecture originally described by the authors. However, upon looking at the results, there is no comparison that can be drawn from the results of the two networks. Since GANs are difficult to anticipate just by the loss number, the comparison cannot be made. The results are pretty comparable for the Bayesian models and the original DCGAN architecture. 

\begin{figure}[H]
\begin{center}
\includegraphics[height=.7\textheight]{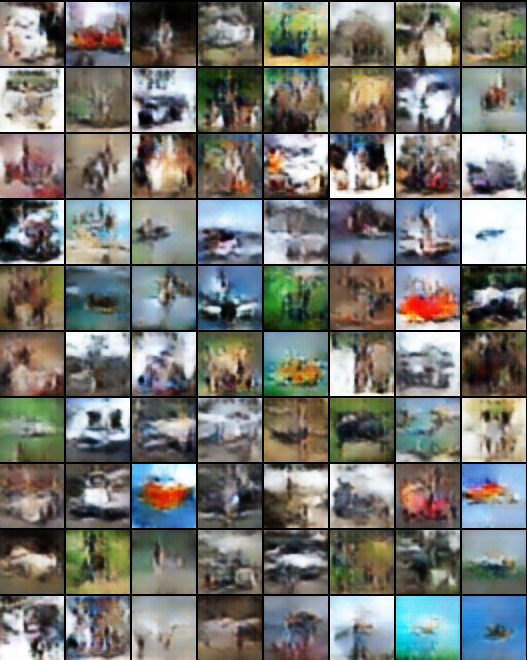}
\label{fig:FakeSamples}
\caption{Fake Samples generated from the Bayesian DCGAN model trained on CIFAR10 dataset}
\end{center}
\end{figure}

\begin{figure}[H]
\begin{center}
\includegraphics[height=.7\textheight]{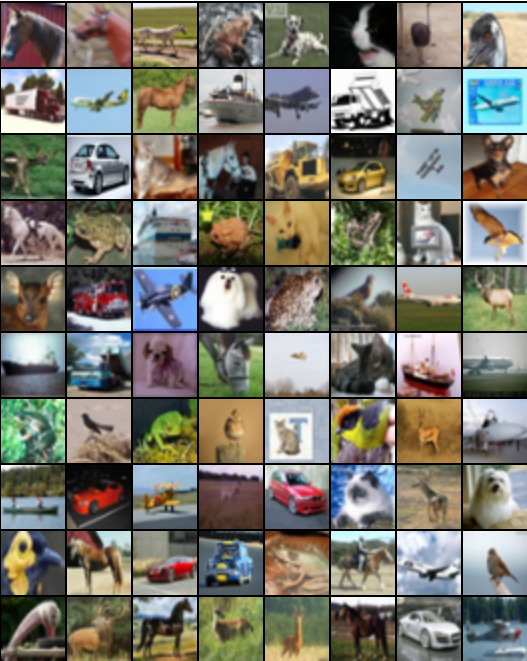}
\label{fig:RealSamples}
\caption{Real Samples taken from CIFAR10 dataset}
\end{center}
\end{figure}

\section{Conclusion and Outlook}

We propose Bayesian \acp{cnn} utilizing \textit{Bayes by Backprop} as a reliable, variational inference method for \acp{cnn} which has not been studied to-date, and estimate the models' aleatoric and epistemic uncertainties for prediction. Furthermore, we apply different ways to pruning the Bayesian \ac{cnn} and compare its results with frequentist architectures.

There has been previous work by Gal and Ghahramani \cite{gal2015bayesian} who utilized the various outputs of a Dropout function to define a distribution, and concluded that one can then speak of a Bayesian \ac{cnn}. This approach finds, perhaps also due to its ease, a large confirming audience. However, we argue against this approach and claim deficiencies. Specifically, in Gal's and Ghahramani's \cite{gal2015bayesian} approach, no prior probability distributions $p(w)$ are placed on the \ac{cnn}'s parameters. But, these are a substantial part of a Bayesian interpretation for the simple reason that Bayes' theorem includes them. Thus we argue, starting with prior probability distributions $p(w)$ is essential in Bayesian methods. In comparison, we place prior probability distributions over all model parameters and update them according to Bayes' theorem with variational inference, precisely \textit{Bayes by Backprop}. We show that these neural networks achieve state-of-the-art results as those achieved by the same network architectures trained by frequentist inference.

Furthermore, we examine how uncertainties (both aleatoric and epistemic uncertainties) can be computed for our proposed method and we show how epistemic uncertainties can be reduced upon more training data. We also compare the effect of dropout in a frequentist network to the proposed Bayesian \ac{cnn} and show the natural regularization effect of Bayesian methods. To counter the twice number of parameters (mean and variance) in a Bayesian \ac{cnn} compared to a single point estimate weight in a frequentist method, we apply methods of network pruning and show that the Bayesian \ac{cnn} performs equally good or better even when the network is pruned and the number of parameters is made comparable to a frequentist method.

Finally, we show the applications of Bayesian \acp{cnn} in various domains like Image recognition, Image Super-Resolution and Generative Adversarial Networks (GANs). The results are compared with other popular approaches in the field and a comparison of results are drawn. Bayesian \acp{cnn} in general, proved to be a good idea to be applied on GANs as prior knowledge for discriminator network helps in better identification of real vs fake images. \\

As an add-on method to further enhance the stability of the optimization, \textit{posterior sharpening} \cite{fortunato2017bayesian} could be applied to Bayesian \acp{cnn} in future work. There, the variational posterior distribution $q_{\theta}(w|\mathcal{D})$ is conditioned on the training data of a batch $\mathcal{D}^{(i)}$. We can see $q_{\theta}(w|\mathcal{D}^{(i)})$ as a proposal distribution, or \textit{hyper-prior} when we rethink it as a hierarchical model, to improve the gradient estimates of the intractable likelihood function $p(\mathcal{D}|w)$. For the initialization of the mean and variance, a zero mean and one as standard deviation was used as the normal distribution seems to be the most intuitive distribution to start with. However, with the results drawn in the thesis from several experimentations, a zero-centred mean and very small standard deviation initialization seemed to be performing equally well but training faster. Xavier initialization \cite{glorot2010understanding} converges faster in a frequentist network compared to a normal initialization and a similar distribution space needs to be explored with Bayesian networks for initializing the distribution. Other properties like periodicity or spatial invariance are also captured by the priors in data space, and based on these properties an alternative to Gaussian process priors can be found. 

Using normal distribution as prior for uncertainty estimation was also explored by Danijar et al. \cite{hafner2018reliable} and it was observed that standard normal prior causes the function posterior to generalize in unforeseen ways on inputs outside of the training distribution. Addition of some noise in the normal distribution as prior can help in better uncertainty estimation by the model. However, no such cases were found in our experiments but can be an interesting area to explore in future.

The network is pruned with simple methods like L1 norm and more compression tricks like vector quantization \cite{DBLP:journals/corr/GongLYB14} and group sparsity regularization  \cite{DBLP:conf/nips/AlvarezS16} can be applied. In our work, we show that reducing the number of model parameters results in a better generalization of the Bayesian architecture and even leads to improvement in the overall model accuracy on the test dataset. Upon further analysis of the model, there is no concrete learning about the change in the behaviour. A more detailed analysis by visualizing the pattern learned by each neuron and grouping them together and removing the redundant neurons which learns similar behaviour is a good way to prune the model.

The concept of Bayesian \ac{cnn} is applied to the discriminative network of a GAN in our work and it has shown good initial results. However, the area of Bayesian generative networks in a GAN is still to be investigated.

\bibliographystyle{plain}
\bibliography{references}

\begin{thebibliography}{10}

\bibitem{DBLP:conf/nips/AlvarezS16}
Jose~M. Alvarez and Mathieu Salzmann.
\newblock Learning the number of neurons in deep networks.
\newblock In {\em {NIPS}}, pages 2262--2270, 2016.

\bibitem{DBLP:journals/corr/AnwarHS15}
Sajid Anwar, Kyuyeon Hwang, and Wonyong Sung.
\newblock Structured pruning of deep convolutional neural networks.
\newblock {\em CoRR}, abs/1512.08571, 2015.

\bibitem{barber1998ensemble}
David Barber and Christopher~M Bishop.
\newblock Ensemble learning in bayesian neural networks.
\newblock {\em NATO ASI SERIES F COMPUTER AND SYSTEMS SCIENCES}, 168:215--238,
  1998.

\bibitem{blundell2015weight}
Charles Blundell, Julien Cornebise, Koray Kavukcuoglu, and Daan Wierstra.
\newblock Weight uncertainty in neural networks.
\newblock {\em arXiv preprint arXiv:1505.05424}, 2015.

\bibitem{buntine1991bayesian}
Wray~L Buntine and Andreas~S Weigend.
\newblock Bayesian back-propagation.
\newblock {\em Complex systems}, 5(6):603--643, 1991.

\bibitem{DBLP:journals/corr/ChangpinyoSZ17}
Soravit Changpinyo, Mark Sandler, and Andrey Zhmoginov.
\newblock The power of sparsity in convolutional neural networks.
\newblock {\em CoRR}, abs/1702.06257, 2017.

\bibitem{denker1991transforming}
John~S Denker and Yann LeCu.
\newblock Transforming neural-net output levels to probability distributions.
\newblock In {\em Advances in neural information processing systems}, pages
  853--859, 1991.

\bibitem{denton2014exploiting}
Emily~L Denton, Wojciech Zaremba, Joan Bruna, Yann LeCun, and Rob Fergus.
\newblock Exploiting linear structure within convolutional networks for
  efficient evaluation.
\newblock In {\em Advances in neural information processing systems}, pages
  1269--1277, 2014.

\bibitem{Kiureghian}
Armen Der~Kiureghian and Ove Ditlevsen.
\newblock Aleatory or epistemic? does it matter?
\newblock {\em Structural Safety}, 31:105--112, 03 2009.

\bibitem{Rumelhart}
David E.~Rumelhart, Geoffrey E.~Hinton, and Ronald J.~Williams.
\newblock Learning representations by back propagating errors.
\newblock {\em Nature}, 323:533--536, 10 1986.

\bibitem{fortunato2017bayesian}
Meire Fortunato, Charles Blundell, and Oriol Vinyals.
\newblock Bayesian recurrent neural networks.
\newblock {\em arXiv preprint arXiv:1704.02798}, 2017.

\bibitem{friston2007variational}
Karl Friston, J{\'e}r{\'e}mie Mattout, Nelson Trujillo-Barreto, John Ashburner,
  and Will Penny.
\newblock Variational free energy and the laplace approximation.
\newblock {\em Neuroimage}, 34(1):220--234, 2007.

\bibitem{gal2015bayesian}
Yarin Gal and Zoubin Ghahramani.
\newblock Bayesian convolutional neural networks with bernoulli approximate
  variational inference.
\newblock {\em arXiv preprint arXiv:1506.02158}, 2015.

\bibitem{Gal2015Dropout}
Yarin Gal and Zoubin Ghahramani.
\newblock Dropout as a {B}ayesian approximation: Insights and applications.
\newblock In {\em Deep Learning Workshop, ICML}, 2015.

\bibitem{glorot2010understanding}
Xavier Glorot and Yoshua Bengio.
\newblock Understanding the difficulty of training deep feedforward neural
  networks.
\newblock In {\em Proceedings of the thirteenth international conference on
  artificial intelligence and statistics}, pages 249--256, 2010.

\bibitem{Gluon}
{Gluon MXnet}.
\newblock chapter18\_variational-methods-and-uncertainty.
\newblock
  \url{https://gluon.mxnet.io/chapter18_variational-methods-and-uncertainty/bayes-by-backprop.html},
  2017.
\newblock Online.

\bibitem{DBLP:journals/corr/GongLYB14}
Yunchao Gong, Liu Liu, Ming Yang, and Lubomir~D. Bourdev.
\newblock Compressing deep convolutional networks using vector quantization.
\newblock {\em CoRR}, abs/1412.6115, 2014.

\bibitem{goodfellow2014generative}
Ian Goodfellow, Jean Pouget-Abadie, Mehdi Mirza, Bing Xu, David Warde-Farley,
  Sherjil Ozair, Aaron Courville, and Yoshua Bengio.
\newblock Generative adversarial nets.
\newblock In {\em Advances in neural information processing systems}, pages
  2672--2680, 2014.

\bibitem{graves2011practical}
Alex Graves.
\newblock Practical variational inference for neural networks.
\newblock In {\em Advances in Neural Information Processing Systems}, pages
  2348--2356, 2011.

\bibitem{graves2016stochastic}
Alex Graves.
\newblock Stochastic backpropagation through mixture density distributions.
\newblock {\em arXiv preprint arXiv:1607.05690}, 2016.

\bibitem{1203152}
B.~K. Gunturk, A.~U. Batur, Y.~Altunbasak, M.~H. Hayes, and R.~M. Mersereau.
\newblock Eigenface-domain super-resolution for face recognition.
\newblock {\em IEEE Transactions on Image Processing}, 12(5):597--606, May
  2003.

\bibitem{hafner2018reliable}
Danijar Hafner, Dustin Tran, Alex Irpan, Timothy Lillicrap, and James Davidson.
\newblock Reliable uncertainty estimates in deep neural networks using noise
  contrastive priors.
\newblock {\em arXiv preprint arXiv:1807.09289}, 2018.

\bibitem{DBLP:journals/corr/HanMD15}
Song Han, Huizi Mao, and William~J. Dally.
\newblock Deep compression: Compressing deep neural network with pruning,
  trained quantization and huffman coding.
\newblock {\em CoRR}, abs/1510.00149, 2015.

\bibitem{he2016deep}
Kaiming He, Xiangyu Zhang, Shaoqing Ren, and Jian Sun.
\newblock Deep residual learning for image recognition.
\newblock In {\em Proceedings of the IEEE conference on computer vision and
  pattern recognition}, pages 770--778, 2016.

\bibitem{hinton2012improving}
Geoffrey~E Hinton, Nitish Srivastava, Alex Krizhevsky, Ilya Sutskever, and
  Ruslan~R Salakhutdinov.
\newblock Improving neural networks by preventing co-adaptation of feature
  detectors.
\newblock {\em arXiv preprint arXiv:1207.0580}, 2012.

\bibitem{hinton1993keeping}
Geoffrey~E Hinton and Drew Van~Camp.
\newblock Keeping the neural networks simple by minimizing the description
  length of the weights.
\newblock In {\em Proceedings of the sixth annual conference on Computational
  learning theory}, pages 5--13. ACM, 1993.

\bibitem{hochreiter1995simplifying}
Sepp Hochreiter and J{\"u}rgen Schmidhuber.
\newblock Simplifying neural nets by discovering flat minima.
\newblock In {\em Advances in neural information processing systems}, pages
  529--536, 1995.

\bibitem{houthooft2016curiosity}
Rein Houthooft, Xi~Chen, Yan Duan, John Schulman, Filip De~Turck, and Pieter
  Abbeel.
\newblock Curiosity-driven exploration in deep reinforcement learning via
  bayesian neural networks.
\newblock {\em arXiv preprint arxiv.1605.09674}, 2016.

\bibitem{karparthy}
{Karparthy, Andrej}.
\newblock {Neural Networks 1}.
\newblock \url{http://cs231n.github.io/neural-networks-1/}, 2016.
\newblock Online.

\bibitem{kendall2017uncertainties}
Alex Kendall and Yarin Gal.
\newblock What uncertainties do we need in bayesian deep learning for computer
  vision?
\newblock In {\em Advances in neural information processing systems}, pages
  5574--5584, 2017.

\bibitem{kingma2014adam}
Diederik~P Kingma and Jimmy Ba.
\newblock Adam: A method for stochastic optimization.
\newblock {\em arXiv preprint arXiv:1412.6980}, 2014.

\bibitem{kingma2015variational}
Diederik~P Kingma, Tim Salimans, and Max Welling.
\newblock Variational dropout and the local reparameterization trick.
\newblock In {\em Advances in Neural Information Processing Systems}, pages
  2575--2583, 2015.

\bibitem{krizhevsky2009learning}
Alex Krizhevsky and Geoffrey Hinton.
\newblock Learning multiple layers of features from tiny images.
\newblock Technical report, Citeseer, 2009.

\bibitem{krizhevsky2012imagenet}
Alex Krizhevsky, Ilya Sutskever, and Geoffrey~E Hinton.
\newblock Imagenet classification with deep convolutional neural networks.
\newblock In {\em Advances in neural information processing systems}, pages
  1097--1105, 2012.

\bibitem{kullback1951information}
Solomon Kullback and Richard~A Leibler.
\newblock On information and sufficiency.
\newblock {\em The annals of mathematical statistics}, 22(1):79--86, 1951.

\bibitem{kwon2018uncertainty}
Yongchan Kwon, Joong-Ho Won, Beom~Joon Kim, and Myunghee~Cho Paik.
\newblock Uncertainty quantification using bayesian neural networks in
  classification: Application to ischemic stroke lesion segmentation.
\newblock 2018.

\bibitem{lafferty2001conditional}
John Lafferty, Andrew McCallum, and Fernando~CN Pereira.
\newblock Conditional random fields: Probabilistic models for segmenting and
  labeling sequence data.
\newblock 2001.

\bibitem{DBLP:journals/corr/LebedevGROL14}
Vadim Lebedev, Yaroslav Ganin, Maksim Rakhuba, Ivan~V. Oseledets, and Victor~S.
  Lempitsky.
\newblock Speeding-up convolutional neural networks using fine-tuned
  cp-decomposition.
\newblock {\em CoRR}, abs/1412.6553, 2014.

\bibitem{lecun1998gradient}
Yann LeCun, Leon Bottou, Yoshua Bengio, and Patrick Haffner.
\newblock Gradient-based learning applied to document recognition.
\newblock {\em Proceedings of the IEEE}, 86(11):2278--2324, 1998.

\bibitem{lecun-mnisthandwrittendigit-2010}
Yann LeCun and Corinna Cortes.
\newblock {MNIST} handwritten digit database.
\newblock 2010.

\bibitem{lecun1990optimal}
Yann LeCun, John~S Denker, and Sara~A Solla.
\newblock Optimal brain damage.
\newblock In {\em Advances in neural information processing systems}, pages
  598--605, 1990.

\bibitem{lipton2016efficient}
Zachary~C Lipton, Jianfeng Gao, Lihong Li, Xiujun Li, Faisal Ahmed, and
  Li~Deng.
\newblock Efficient exploration for dialogue policy learning with bbq networks
  \& replay buffer spiking.
\newblock {\em arXiv preprint arXiv:1608.05081}, 2016.

\bibitem{Mackay1991APB}
David J~C Mackay.
\newblock A practical bayesian framework for backprop networks.
\newblock 1991.

\bibitem{mackay1995probable}
David~JC MacKay.
\newblock Probable networks and plausible predictions—a review of practical
  bayesian methods for supervised neural networks.
\newblock {\em Network: Computation in Neural Systems}, 6(3):469--505, 1995.

\bibitem{mackay1996hyperparameters}
David~JC MacKay.
\newblock Hyperparameters: optimize, or integrate out?
\newblock In {\em Maximum entropy and bayesian methods}, pages 43--59.
  Springer, 1996.

\bibitem{MartinFTM01}
D.~Martin, C.~Fowlkes, D.~Tal, and J.~Malik.
\newblock A database of human segmented natural images and its application to
  evaluating segmentation algorithms and measuring ecological statistics.
\newblock In {\em Proc. 8th Int'l Conf. Computer Vision}, volume~2, pages
  416--423, July 2001.

\bibitem{molchanov2017variational}
Dmitry Molchanov, Arsenii Ashukha, and Dmitry Vetrov.
\newblock Variational dropout sparsifies deep neural networks.
\newblock {\em arXiv preprint arXiv:1701.05369}, 2017.

\bibitem{DBLP:journals/corr/NarangDSE17}
Sharan Narang, Gregory~F. Diamos, Shubho Sengupta, and Erich Elsen.
\newblock Exploring sparsity in recurrent neural networks.
\newblock {\em CoRR}, abs/1704.05119, 2017.

\bibitem{neal2012bayesian}
Radford~M Neal.
\newblock {\em Bayesian learning for neural networks}, volume 118.
\newblock Springer Science \& Business Media, 2012.

\bibitem{neal1998view}
Radford~M Neal and Geoffrey~E Hinton.
\newblock A view of the em algorithm that justifies incremental, sparse, and
  other variants.
\newblock In {\em Learning in graphical models}, pages 355--368. Springer,
  1998.

\bibitem{neklyudov2018variance}
Kirill Neklyudov, Dmitry Molchanov, Arsenii Ashukha, and Dmitry Vetrov.
\newblock Variance networks: When expectation does not meet your expectations.
\newblock {\em arXiv preprint arXiv:1803.03764}, 2018.

\bibitem{DBLP:journals/corr/RadfordMC15}
Alec Radford, Luke Metz, and Soumith Chintala.
\newblock Unsupervised representation learning with deep convolutional
  generative adversarial networks.
\newblock {\em CoRR}, abs/1511.06434, 2015.

\bibitem{DBLP:journals/corr/ShiCHTABRW16}
Wenzhe Shi, Jose Caballero, Ferenc Husz{\'{a}}r, Johannes Totz, Andrew~P.
  Aitken, Rob Bishop, Daniel Rueckert, and Zehan Wang.
\newblock Real-time single image and video super-resolution using an efficient
  sub-pixel convolutional neural network.
\newblock {\em CoRR}, abs/1609.05158, 2016.

\bibitem{10.1007/978-3-642-40760-4_2}
Wenzhe Shi, Jose Caballero, Christian Ledig, Xiahai Zhuang, Wenjia Bai, Kanwal
  Bhatia, Antonio M. Simoes~Monteiro de~Marvao, Tim Dawes, Declan O'Regan, and
  Daniel Rueckert.
\newblock Cardiac image super-resolution with global correspondence using
  multi-atlas patchmatch.
\newblock In Kensaku Mori, Ichiro Sakuma, Yoshinobu Sato, Christian Barillot,
  and Nassir Navab, editors, {\em Medical Image Computing and Computer-Assisted
  Intervention -- MICCAI 2013}. Springer Berlin Heidelberg, 2013.

\bibitem{shridhar2018bayesian}
Kumar Shridhar, Felix Laumann, Adrian Llopart~Maurin, Martin Olsen, and Marcus
  Liwicki.
\newblock Bayesian convolutional neural networks with variational inference.
\newblock {\em arXiv preprint arXiv:1806.05978}, 2018.

\bibitem{simonyan2014very}
Karen Simonyan and Andrew Zisserman.
\newblock Very deep convolutional networks for large-scale image recognition.
\newblock {\em arXiv preprint arXiv:1409.1556}, 2014.

\bibitem{Soudry:NIPS2014_5269}
Daniel Soudry, Itay Hubara, and Ron Meir.
\newblock Expectation backpropagation: Parameter-free training of multilayer
  neural networks with continuous or discrete weights.
\newblock In Z.~Ghahramani, M.~Welling, C.~Cortes, N.~D. Lawrence, and K.~Q.
  Weinberger, editors, {\em Advances in Neural Information Processing Systems
  27}, pages 963--971. Curran Associates, Inc., 2014.

\bibitem{srivastava2014dropout}
Nitish Srivastava, Geoffrey Hinton, Alex Krizhevsky, Ilya Sutskever, and Ruslan
  Salakhutdinov.
\newblock Dropout: a simple way to prevent neural networks from overfitting.
\newblock {\em The Journal of Machine Learning Research}, 15(1):1929--1958,
  2014.

\bibitem{tibshirani1996regression}
Robert Tibshirani.
\newblock Regression shrinkage and selection via the lasso.
\newblock {\em Journal of the Royal Statistical Society. Series B
  (Methodological)}, pages 267--288, 1996.

\bibitem{Torralba:2008:MTI:1444381.1444403}
Antonio Torralba, Rob Fergus, and William~T. Freeman.
\newblock 80 million tiny images: A large data set for nonparametric object and
  scene recognition.
\newblock {\em IEEE Trans. Pattern Anal. Mach. Intell.}, 30(11), November 2008.

\bibitem{wang2013fast}
Sida Wang and Christopher Manning.
\newblock Fast dropout training.
\newblock In {\em international conference on machine learning}, pages
  118--126, 2013.

\bibitem{Yang2014SingleImageSA}
Chih-Yuan Yang, Chao Ma, and Ming-Hsuan Yang.
\newblock Single-image super-resolution: A benchmark.
\newblock In {\em ECCV}, 2014.

\bibitem{yedidia2005constructing}
Jonathan~S Yedidia, William~T Freeman, and Yair Weiss.
\newblock Constructing free-energy approximations and generalized belief
  propagation algorithms.
\newblock {\em IEEE Transactions on information theory}, 51(7):2282--2312,
  2005.

\end{thebibliography}

\section{Appendix}

\section{Experiment Specifications}

\section*{Bayesian Settings}

\subsection{Image Recognition}

\begin{table}[H]
    \centering
    \renewcommand{\arraystretch}{2}
    \begin{tabular}[c]{c | c} 
     \hline
     variable & value \\ [0.5ex] 
     \hline
     learning rate &  0.001\\ 
     
     epochs & 100 \\
     
     batch size & 256 \\
     
     sample size & 10-25 \\
     
     loss & cross-entropy \\
     
     $(\alpha \mu^2)_{init}$ of approximate posterior $q_{\theta}(w|\mathcal{D})$ & -10 \\
     
     optimizer & Adam \cite{kingma2014adam} \\
     
     $\lambda$ in $\ell$-2 normalisation & 0.0005 \\
    
     $\beta_i$ & $\frac{2^{M-i}}{2^M-1}$ \cite{blundell2015weight} \\ [1ex] 
     \hline
    \end{tabular} 
    \renewcommand{\arraystretch}{2}
\end{table}

Sample size can vary from 10 to 25 as this range provided the best results. However, it can be played around with. For most of our experiments, it is either 10 or 25 unless specified otherwise. 

\subsection{Image Super Resolution}

\begin{table}[H]
    \centering
    \renewcommand{\arraystretch}{2}
    \begin{tabular}[c]{c | c} 
     \hline
     variable & value \\ [0.5ex] 
     \hline
     learning rate &  0.01\\ 
     
     epochs & 200 \\
     
     batch size & 64 \\
     
     upscale factor & 3 \\
     
     loss & Mean Squared Error \\
     
     seed & 123 \\
     
     $(\alpha \mu^2)_{init}$ of approximate posterior $q_{\theta}(w|\mathcal{D})$ & -10 \\
     
     optimizer & Adam \cite{kingma2014adam} \\
     
     $\lambda$ in $\ell$-2 normalisation & 0.0005 \\
    
     $\beta_i$ & $\frac{2^{M-i}}{2^M-1}$ \cite{blundell2015weight} \\ [1ex] 
     \hline
    \end{tabular} 
    \renewcommand{\arraystretch}{2}
\end{table}

\subsection{Generative Adversarial Network}

\begin{table}[H]
    \centering
    \renewcommand{\arraystretch}{2}
    \begin{tabular}[c]{c | c} 
     \hline
     variable & value \\ [0.5ex] 
     \hline
     learning rate &  0.001\\ 
     
     epochs & 100 \\
     
     batch size & 64 \\
     
     image size & 64 \\
     
     latent vector (nz) & 100 \\
     
     number of generator factor (ndf) & 64 \\
     
     number of discriminator factor (ndf) & 64 \\
     
     upscale factor & 3 \\
     
     loss & Mean Squared Error \\
     
     number of channels (nc) & 3 \\
     
     $(\alpha \mu^2)_{init}$ of approximate posterior $q_{\theta}(w|\mathcal{D})$ & -10 \\
     
     optimizer & Adam \cite{kingma2014adam} \\
     
     $\lambda$ in $\ell$-2 normalisation & 0.0005 \\
    
     $\beta_i$ & $\frac{2^{M-i}}{2^M-1}$ \cite{blundell2015weight} \\ [1ex] 
     \hline
    \end{tabular} 
    \renewcommand{\arraystretch}{2}
\end{table}

\section*{Non Bayesian Settings}

\subsection{Image Recognition}

\begin{table}[H]
    \centering
    \renewcommand{\arraystretch}{2}
    \begin{tabular}[c]{c | c} 
     \hline
     variable & value \\ [0.5ex] 
     \hline
     learning rate &  0.001\\ 
     
     epochs & 100 \\
     
     batch size & 256 \\
     
     loss & cross-entropy \\
     
     initializer & Xavier \cite{glorot2010understanding} or Normal \\
     
     optimizer & Adam \cite{kingma2014adam} \\ [1ex] 
     \hline
    \end{tabular} 
    \renewcommand{\arraystretch}{2}
\end{table}

The weights were initialized with Xavier initialization \cite{glorot2010understanding} at first, but to make it consistent with the Bayesian networks where initialization was Normal initialization (mean = 0 and variance = 1), the initializer was changed to Normal initialization.

\pagebreak

\section*{Architectures}

\subsection{LeNet-5}
 
\begin{table}[H]
    \centering
    \renewcommand{\arraystretch}{2}
    \begin{tabular}{c c c c c c} 
     \hline
     layer type & width & stride & padding & input shape & nonlinearity \\ [0.5ex] 
     \hline
     convolution ($5\times5$) & 6 & 1 & 0 & $M\times1\times32\times32$ & Softplus \\ 
     
     Mmax-pooling ($2\times2$) & \empty & 2 & 0 & $M\times6\times28\times28$ & \empty \\
     
     convolution ($5\times5$) & 16 & 1 & 0 & $M\times1\times14\times14$ & Softplus \\
     
     max-pooling ($2\times2$) & \empty & 2 & 0 & $M\times16\times10\times10$ & \empty \\
    
     fully-connected & 120 & \empty & \empty & $M\times400$ & Softplus \\
     
     fully-connected & 84 & \empty & \empty & $M\times120$ & Softplus \\
     
     fully-connected & 10 & \empty & \empty & $M\times84$ & \empty \\ [1ex] 
     \hline
    \end{tabular} 
    \renewcommand{\arraystretch}{1}
    \label{tab:LeNet}
    \caption{LeNet architecture with original configurations as defined in the paper. \cite{lecun1998gradient}}
\label{tab:AlexNet}

\end{table}

\subsection{AlexNet}

\begin{table}[H]
    \centering
    \renewcommand{\arraystretch}{2}
    \begin{tabular}{c c c c c c} 
 \hline
 layer type & width & stride & padding & input shape & nonlinearity \\ [0.5ex] 
 \hline
 convolution ($11\times11$) & 64 & 4 & 5 & $M\times3\times32\times32$ & Softplus \\ 
 
 max-pooling ($2\times2$) & \empty & 2 & 0 & $M\times64\times32\times32$ & \empty \\
 
 convolution ($5\times5$) & 192 & 1 & 2 & $M\times64\times15\times15$ & Softplus \\
 
 max-pooling ($2\times2$) & \empty & 2 & 0 & $M\times192\times15\times15$ & \empty \\
 
 convolution ($3\times3$) & 384 & 1 & 1 & $M\times192\times7\times7$ & Softplus \\
 
 convolution ($3\times3$) & 256 & 1 & 1 & $M\times384\times7\times7$ & Softplus \\
 
 convolution ($3\times3$) & 128 & 1 & 1 & $M\times256\times7\times7$ & Softplus \\
 
 max-pooling ($2\times2$) & \empty & 2 & 0 & $M\times128\times7\times7$ & \empty \\
 
 fully-connected & 128 & \empty & \empty & $M\times128$ & \empty \\ [1ex] 
 \hline
\end{tabular}
\renewcommand{\arraystretch}{1}
\caption{AlexNet architecture with original configurations as defined in the paper. \cite{krizhevsky2012imagenet}}
\label{tab:AlexNet}
\end{table}

\subsection{AlexNetHalf}

\begin{table}[H]
    \centering
    \renewcommand{\arraystretch}{2}
    \begin{tabular}{c c c c c c} 
 \hline
 layer type & width & stride & padding & input shape & nonlinearity \\ [0.5ex] 
 \hline
 convolution ($11\times11$) & 32 & 4 & 5 & $M\times3\times32\times32$ & Softplus \\ 
 
 max-pooling ($2\times2$) & \empty & 2 & 0 & $M\times32\times32\times32$ & \empty \\
 
 convolution ($5\times5$) & 96 & 1 & 2 & $M\times32\times15\times15$ & Softplus \\
 
 max-pooling ($2\times2$) & \empty & 2 & 0 & $M\times96\times15\times15$ & \empty \\
 
 convolution ($3\times3$) & 192 & 1 & 1 & $M\times96\times7\times7$ & Softplus \\
 
 convolution ($3\times3$) & 128 & 1 & 1 & $M\times192\times7\times7$ & Softplus \\
 
 convolution ($3\times3$) & 64 & 1 & 1 & $M\times128\times7\times7$ & Softplus \\
 
 max-pooling ($2\times2$) & \empty & 2 & 0 & $M\times64\times7\times7$ & \empty \\
 
 fully-connected & 64 & \empty & \empty & $M\times64$ & \empty \\ [1ex] 
 \hline
\end{tabular}
\renewcommand{\arraystretch}{1.5}
\label{tab:AlexNetHalfArchitecture}
\caption{AlexNetHalf with number of filters halved compared to the original architecture.}
\end{table}

\section{How to replicate results}

Install PyTorch from the official website (\url{https://pytorch.org/})

\begin{verbatim} 

git clone https://github.com/kumar-shridhar/PyTorch-BayesianCNN
pip install -r requirements.txt

\end{verbatim}
\textit{cd} into respective folder/ task to replicate (Image Recognition, Super Resolution or GAN)

\subsection{Image Recognition}

\begin{verbatim} 
python main_Bayesian.py
\end{verbatim}

to replicate the Bayesian \acp{cnn} results.

\begin{verbatim} 
python main_nonBayesian.py
\end{verbatim}

to replicate the Frequentist \acp{cnn} results.\\

For more details, read the README sections of the repo : \url{https://github.com/kumar-shridhar/PyTorch-BayesianCNN}

\end{document}